\documentclass[runningheads]{llncs}

\usepackage{eccv}

\usepackage{eccvabbrv}

\usepackage{graphicx}
\usepackage{overpic}
\usepackage{booktabs}

\usepackage[accsupp]{axessibility}  % Improves PDF readability for those with disabilities.
\newcommand{\hu}[1]{\textcolor{red}{#1}}
\renewcommand{\hu}{}
\usepackage{hyperref}

\usepackage{orcidlink}

\begin{document}

\title{Unsupervised Exposure Correction} 


\author{Ruodai Cui\inst{1} \and
Li Niu\inst{2}\thanks{Corresponding author.} \and
Guosheng Hu\inst{3}}

\authorrunning{Ruodai et al.}

\institute{Qualcomm Technologies, Inc.\\ \email{ruodcui@qti.qualcomm.com}\and
Department of Computer Science and Engineering, MoE Key Lab of Artificial Intelligence, Shanghai Jiao Tong University \\
\email{ustcnewly@sjtu.edu.cn}\\ \and
University of Bristol}

\maketitle

\begin{abstract}
Current exposure correction methods have three challenges, labor-intensive paired data annotation, limited generalizability, and performance degradation in low-level computer vision tasks. In this work, we introduce an innovative Unsupervised Exposure Correction (UEC) method that eliminates the need for manual annotations, offers improved generalizability, and enhances performance in low-level downstream tasks. Our model is trained using \emph{freely} available paired data from an emulated Image Signal Processing (ISP) pipeline. This approach does not need expensive manual annotations, thereby minimizing individual style biases from the annotation and consequently improving its generalizability. Furthermore, we present a large-scale Radiometry Correction Dataset, specifically designed to emphasize exposure variations, to facilitate unsupervised learning. In addition, we develop a transformation function that preserves image details and outperforms state-of-the-art supervised methods\cite{Eyiokur2022}, while utilizing only 0.01\% of their parameters. Our work further investigates the broader impact of exposure correction on downstream tasks, including edge detection, demonstrating its effectiveness in mitigating the adverse effects of poor exposure on low-level features. The source code and dataset are publicly available at \url{https://github.com/BeyondHeaven/uec_code}.

\end{abstract}

\section{Introduction}
\label{sec:intro}

% polish the follwing content, wording of your result should not contain superfluous words that are inappropriate for a scientific article

% When performing in auto-exposure mode, exposure values will be adjusted to compensate for low/high levels of brightness in the captured environment by metering through-the-lens, which means the amount of light received from the scene. Exposure errors can occur due to several factors, such as errors in measurements of through-the-lens metering, dramatic changes of the brightness level in the scene, hard lighting conditions (e.g. , very low lighting and backlighting), and errors made by users in manual mode.

Exposure, a pivotal factor in photography, significantly impacts image quality by influencing visual clarity.  In radiometry, the exposure is defined as scene irradiance, the amount of light that reaches the image sensor. This can be controlled by exposure value (EV), which combines essential factors such as aperture, shutter speed, and ISO settings. Despite advancements in Image Signal Processor (ISP), which facilitate automatic EV adjustments during image capture, challenges persist under non-ideal lighting conditions. As a result, post-processing of sRGB images remains crucial. The advent of deep learning has spurred extensive research in this field, with numerous studies proposing models for exposure correction \cite{ren2019low, zhang2019kindling, guo2020zero, afifi2020learning, jiang2021enlightengan, li2021learning, liu2021retinex, nsampi2021learning, hu2021contextual, li2022cudi, bhattacharya2022d2bgan, yang2023learning}, demonstrating significant achievements. Nevertheless, these methods still face three challenges.

(1) A primary challenge arises from the dependence on the paired data, where the ground truth is from proficient photographers. This process is inherently complex and labor-intensive, as it involves detailed editing and refinement of each image, demanding more manual work than just labeling as in classification tasks.
(2) Previous methods often suffer from limited generalizability.  To be specific, (a) as noted, the efficiency of manual adjustments is low, resulting in small datasets available in academia. (b) the manual adjustments inevitably introduce different stylistic biases. Personal preferences vary significantly between individuals, suggesting that such ground truths are intrinsically noisy.
(3) Previous methodologies have predominantly focused on generating aesthetically pleasing images. However, they often yield images with notable degradations in low-level features. Such degradations render these images less suitable for various downstream computer vision tasks, like edge detection and segmentation, where preserving these features is important.

\begin{figure}[t]
    % \parbox[t][0.02\linewidth][c]{0.45\linewidth}{\scriptsize \centering (a) Input Multi-exposure Images}
    % \parbox[t][0.02\linewidth][c]{0.45\linewidth}{\scriptsize \centering (b) Ground Truth with Stylistic Biases}
    \begin{subfigure}[t]{0.48\linewidth}
    % \centering
    \begin{overpic}[width=0.18\textwidth,height=0.24\textwidth]{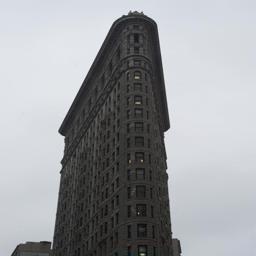}
        \put(2,90){\textcolor[rgb]{1,1,1}{\tiny +1.5EV}}
    \end{overpic}
    \begin{overpic}[width=0.18\textwidth,height=0.24\textwidth]{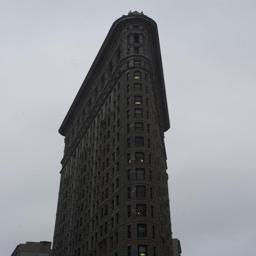}
        \put(2,90){\textcolor[rgb]{1,1,1}{\tiny +1.0EV}}
    \end{overpic}
    \begin{overpic}[width=0.18\textwidth,height=0.24\textwidth]{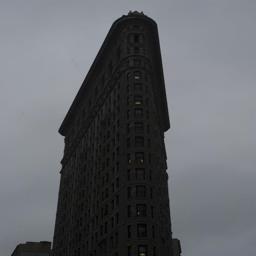}
        \put(2,90){\textcolor[rgb]{1,1,1}{\tiny 0EV}}
    \end{overpic}
    \begin{overpic}[width=0.18\textwidth,height=0.24\textwidth]{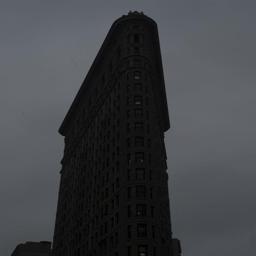}
        \put(2,90){\textcolor[rgb]{1,1,1}{\tiny -1.0EV}}
    \end{overpic}
    \begin{overpic}[width=0.18\textwidth,height=0.24\textwidth]{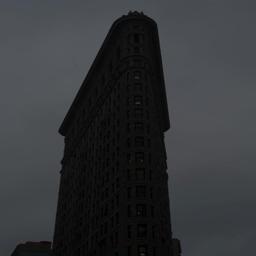}
        \put(2,90){\textcolor[rgb]{1,1,1}{\tiny -1.5EV}}
    \end{overpic}
    \caption{Input Multi-exposure Sequence}
    \label{fig:artifact_input}
    \end{subfigure}
    \begin{subfigure}[t]{0.48\linewidth}
    \begin{overpic}[width=0.18\textwidth,height=0.24\textwidth]{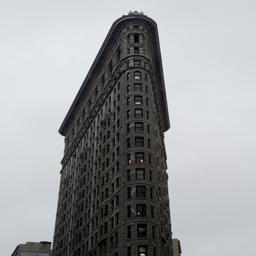}
        \put(2,90){\textcolor[rgb]{1,1,1}{\tiny ExpertA}}
    \end{overpic}
    \begin{overpic}[width=0.18\textwidth,height=0.24\textwidth]{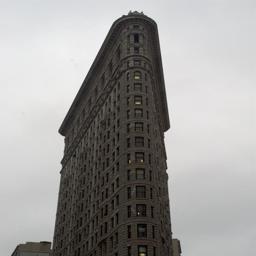}
        \put(2,90){\textcolor[rgb]{1,1,1}{\tiny ExpertB}}
    \end{overpic}
    \begin{overpic}[width=0.18\textwidth,height=0.24\textwidth]{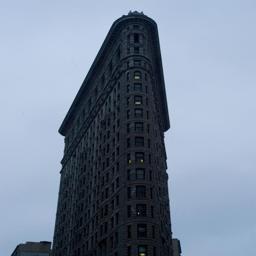}
        \put(2,90){\textcolor[rgb]{1,1,1}{\tiny ExpertC}}
    \end{overpic}
    \begin{overpic}[width=0.18\textwidth,height=0.24\textwidth]{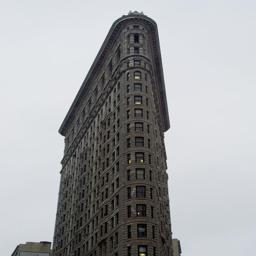}
        \put(2,90){\textcolor[rgb]{1,1,1}{\tiny ExpertD}}
    \end{overpic}
    \begin{overpic}[width=0.18\textwidth,height=0.24\textwidth]{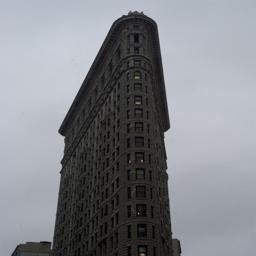}
        \put(2,90){\textcolor[rgb]{1,1,1}{\tiny ExpertE}}
    \end{overpic}
    \caption{Ground Truths with Stylistic Biases}
    \label{fig:artifact_biasGT}
    \end{subfigure}
    \\
    % \parbox[t][0.02\linewidth][c]{0.45\linewidth}{\scriptsize \centering (c) ECM \cite{Eyiokur2022}(Sup.)}
    % \parbox[t][0.02\linewidth][c]{0.45\linewidth}{\scriptsize \centering (d) Ours(Unsup.) }
    \begin{subfigure}[t]{0.48\linewidth}
    \begin{overpic}[width=0.18\textwidth,height=0.24\textwidth]{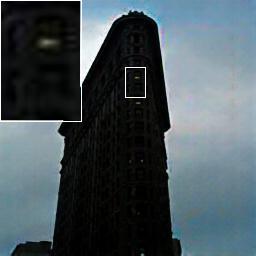}
    \end{overpic}
    \begin{overpic}[width=0.18\textwidth,height=0.24\textwidth]{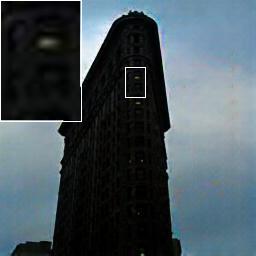}
    \end{overpic}
    \begin{overpic}[width=0.18\textwidth,height=0.24\textwidth]{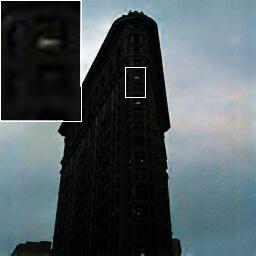}
    \end{overpic}
    \begin{overpic}[width=0.18\textwidth,height=0.24\textwidth]{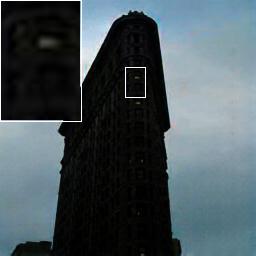}
    \end{overpic}
    \begin{overpic}[width=0.18\textwidth,height=0.24\textwidth]{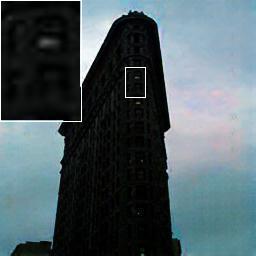}
    \end{overpic}
    \caption{ECM \cite{Eyiokur2022}(Sup.)}
    \label{fig:artifact_ecm}
    \end{subfigure}
    \begin{subfigure}[t]{0.48\linewidth}
    \begin{overpic}[width=0.18\textwidth,height=0.24\textwidth]{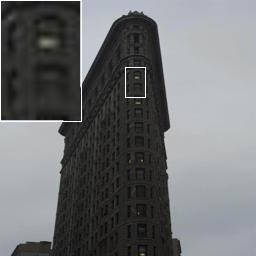}
    \end{overpic}
    \begin{overpic}[width=0.18\textwidth,height=0.24\textwidth]{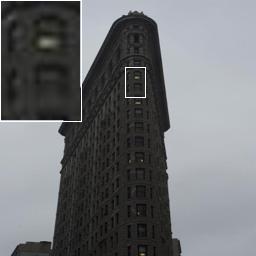}
    \end{overpic}
    \begin{overpic}[width=0.18\textwidth,height=0.24\textwidth]{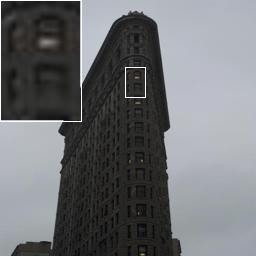}
    \end{overpic}
    \begin{overpic}[width=0.18\textwidth,height=0.24\textwidth]{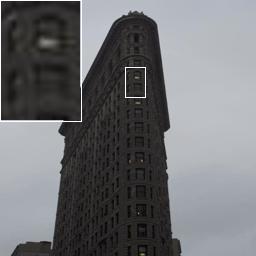}
    \end{overpic}
    \begin{overpic}[width=0.18\textwidth,height=0.24\textwidth]{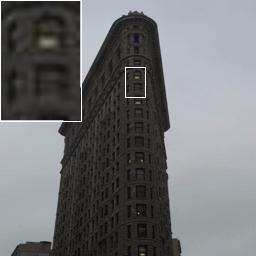}
    \end{overpic}
    \caption{Ours(Unsup.)}
    \label{fig:artifact_ours}
    \end{subfigure}
    \caption{Visual Comparison: ECM \cite{Eyiokur2022} vs. Our UEC method.}
    \label{fig:artifact}
\end{figure}

% having observed that paired images can be freely obtained from the emulated ISP, 

Inspired by Afifi et al.'s work of exposure dataset \cite{Afifi2021}, we realize that acquiring paired data does not necessarily require extensive manual intervention. They apply an emulated Image Signal Processing (ISP) pipeline on RAW data, thereby creating multi-exposure sequences, as shown in Fig.~\ref{fig:artifact_input}. The work \cite{Afifi2021} learns the mapping from the inputs (Fig.~\ref{fig:artifact_input}) to the ground truth (ExpertC in Fig.~\ref{fig:artifact_biasGT}), which is a typical supervised learning.  However, this approach introduces ambiguities due to inconsistencies in the ground truths, e.g. five experts in Fig.~\ref{fig:artifact_biasGT}. Unlike existing supervised learning methods \cite{Afifi2021, Eyiokur2022, guo2020zero}, in this work, we introduce an innovative Unsupervised Exposure Correction (UEC) method.
(1) To achieve this, we creatively ask 
the images in the same multi-exposure sequence, which can be generated freely, to mutually serve as a  ground truth for learning exposure adjustments across a diverse range. Specifically, 
we simply generate multi-exposure sequences and EV labels via an emulated ISP pipeline. 
In this way, we do not need the expensive manual annotations to generate ground truth. 
(2) The existing methods have inconsistent ground truth which might have different standards on colors, leading to degraded performance. In contrast, in our UEC method, the images in multi-exposure sequence do not have such ``standards'' because they mutually work as the ground truth. 
(3) We employ a pixel-wise exposure transformation in sRGB color space for preserving visual details as shown in Fig.~\ref{fig:artifact_ours} which greatly outperforms the SOTA method in Fig.~\ref{fig:artifact_ecm}. 
%which is shown in Fig.~\ref{fig:artifact_ecm} and Fig.~\ref{fig:artifact_ours}.
This detail-preserving property is crucial for downstream computer vision tasks that rely on low-level features, such as edge detection and segmentation.
To further facilitate the analysis without individual stylistic biases in exposure correction, we contribute a new dataset, Radiometry Correction Dataset, that is large and encompasses broad exposures. This dataset maintains a consistent style with only radiometry variations.
%Although these adjustments may not consistently improve image quality, this process greatly reduces the need of manual annotation for final calibration, even a single well-exposed image may suffice. Specifically, our approach involves paired learning for exposure adjustment within the same multi-exposure sequence. It can be further refined by incorporating two reference images from another multi-exposure sequence, where we establish supervision based on preserving the relative brightness in the outputs. 
 
%This methodology offers several advantages:
%(1) Our method is manual annotation free. We simply generate multi-exposure sequences and EV labels via an emulated ISP pipeline. 
%does not need %obviates the need manual data tuning by utilizing the data from an emulated ISP pipeline, including 
%(2) We concentrate on radiometry for better generalizability. Our observations demonstrate that: (a) within any given multi-exposure sequence, radiometry exhibits variability, whereas colorimetry maintains uniformity; (b) across ground truth images from diverse experts, while colorimetry adjustments vary, radiometry consistently remains stable. This focused approach mitigates prevalent issues such as overfitting and stylistic biases, as demonstrated in Fig.~\ref{fig:artifact_biasGT}. 

%while varying only in radiometry, offering a more targeted resource for refining exposure correction techniques.

Our contributions can be summarized as follows:
(1) By modeling radiometry, we introduce an innovative unsupervised learning methodology that fundamentally addresses the problem of expensive annotations. To our knowledge, this is the first unsupervised learning solution for exposure correction, which achieves competitive performance as SOTA supervised models\cite{Eyiokur2022} with only 0.01\% of their parameters, and minimizes individual stylistic biases.
(2) We propose a pixel-wise exposure transformation that can well preserve the details of images and flexibly accept different resolutions of input images. By this way, the enhanced images can improve the performance of many downstream tasks, e.g. edge detection.
(3) We contribute a new dataset  focusing on radiometry for better generalizability. Input images of this dataset   are synthesized from ground truth data, ensuring a uniform style across the dataset.
    
% \begin{enumerate}
%     \item By modeling radiometry, we introduce an innovative unsupervised learning methodology that fundamentally addresses the problem of expensive annotations. To our knowledge, this is the first unsupervised learning solution for exposure correction, which achieves competitive performance as SOTA supervised models\cite{Eyiokur2022} with only 0.01\% of their parameters, and minimizes individual stylistic biases.
%     \item We propose a pixel-wise exposure transformation that can well preserve the details of images and flexibly accept different resolutions of input images. By this way, the enhanced images can improve the performance of many downstream tasks, e.g. edge detection.
%     \item We contribute a new dataset  focusing on radiometry for better generalizability. Input images of this dataset   are synthesized from ground truth data, ensuring a uniform style across the dataset.
% \end{enumerate}

\section{Related Work}

Exposure correction has evolved from traditional techniques to modern deep learning methods. Traditional methods, like histogram equalization \cite{ibrahim2007brightness}, and Retinex-based algorithms \cite{land1971lightness,rahman2004retinex}, laid the groundwork. Recent deep learning studies, including Ren et al. \cite{ren2019low}, Zhang et al. \cite{zhang2019kindling}, Guo et al. \cite{guo2020zero}, and Loh et al. \cite{loh2019getting}, focus on low-light enhancement. However, these methods exposed a gap in addressing both underexposed and overexposed images. Afifi et al. \cite{Afifi2021} introduced a dataset featuring varying exposure attributes, enabling comprehensive methodologies.

Exposure correction methods fall into two categories: image-to-image translation and color transformation. Image-to-image translation aims to create dense translations between input and output pairs, leveraging deep learning as seen in works by Eyiokur et al. \cite{Eyiokur2022}, Afifi et al. \cite{Afifi2021}, and others \cite{chen2018deep,jiang2021enlightengan,bhattacharya2022d2bgan}. Color transformation methods, on the other hand, predict mapping curve parameters to enhance images, employing techniques like quadratic transforms \cite{chai2020supervised,liu2020color,wang2011example,yan2016automatic}, local affine transforms \cite{gharbi2017deep}, curve-based transforms \cite{bychkovsky2011learning,guo2020zero,kim2020global,li2020flexible,moran2021curl}, filters \cite{deng2018aesthetic,moran2020deeplpf}, lookup tables \cite{zeng2020learning,wang2021realtime}, and MLP models \cite{he2020conditional,liu2021very,wang2022neural}. These techniques define specific functions or models for the enhancement process, offering diverse tools for image exposure challenges.

\section{Method}

Traditionally, exposure correction learns a mapping from one input image $I_1$ with one (usually weaker) exposure to another image $I_2$ with a different (usually stronger and better) exposure. To learn this mapping, conventionally, people \emph{manually} generate the well-exposed image $I_2$ as the ground-truth of the existing $I_1$, which is quite expensive and is hard to scale up. 

To address this, we creatively propose a way to exploit the information from the \emph{freely} generated multi-exposure sequence from RAW data. In the same sequence, the images with different exposures can mutually work as the ground-truth to learn the mapping for exposure adjustment. In Sec.~\ref{sec:TheoryModeling}, we formulate our Unsupervised Exposure Correction (UEC). Then, the neural architectures in Sec.~\ref{sec:Model Architecture} and loss functions in Sec.~\ref{sec:Loss Function} are introduced to achieve UEC. Following that, Sec.~\ref{sec:test} outlines the testing methodology. Moreover, we propose our Radiometry Correction Dataset where the  paired images are freely generated from RAW data in Sec.~\ref{sec:Our Dataset}.

%we introduce the free paired images generation from RAW data and propose our Radiometry Correction Dataset in Sec.~\ref{sec:Our Dataset}. 

\subsection{Unsupervised Exposure Correction Modeling} 
\label{sec:TheoryModeling}
The emulated ISP enables us to generate extensive multi-exposure sequences.
%sequences of multi-exposure images. 
To make full use of the large data, we introduce a task with self-supervision. A reference image guides exposure adjustments, serving as the calibration target, while a transformation curve shifts the exposure level. This method involves varying the exposure over a wide range, rather than directly aligning it to an optimal value, thereby categorizing it as a specific form of style transfer.

We initiate the process by employing a style encoder to extract the exposure feature from the image, denoted as \( E = e(I) \). Diverging from conventional style transfer methods, which typically rely on a one-hot vector as a prior \cite{choi2018stargan} or leverage features refined through triplet loss \cite{kotovenko2019content}, we have developed an alternative optimization approach that better accommodates exposure dynamics. We discovered a sequential relationship within varying exposure images, allowing us to order them by their exposures. Consequently, to quantify the difference between any two images, we can employ a single scalar.

\begin{equation}\label{eq:Delta_E}
\Delta E = d(E_1, E_2) = d(e(I_1), e(I_2)). 
\end{equation}

Then, we can perform the style transformation from $\Delta E$:

\begin{equation} \label{eq:I'_1}
I'_1 = f(\Delta E, I_1). 
\end{equation}

Having established the transformation function, , we now focus on developing optimization strategies. We identify two principles to guide the optimization process.

\subsubsection{Restoration Supervision for Pretext Task.} 
Consider a scenario where we sample two images, \(I_1\) and \(I_2\), from the same sequence $I$, leading to the equation \(I'_1 = I_2\). We call this ``Restoration Supervision'', which can train $f(\cdot)$ under the restoration guidance of \(I_2\), as illustrated in Fig.~\ref{fig:modeling1}.

However, it is important to note that both \(I_1\) and \(I_2\) originate from the same sequence $I$. For the real task, we need to handle image pairs from different scenes. Since our method learn from data which varies solely in exposure, the trained $f(\cdot)$ can only modify exposures. Moreover, we calculate the exposure difference in a latent space, not in pixel space, which omits low-level features. This approach enables the difference function $d(\cdot, \cdot)$ to adapt images from different scenes to a certain degree. Therefore, when we select reference images from a different sequence, the alterations remain confined to exposure adjustment, even though the \(\Delta E\) computed may not precisely indicating the optimal exposure variation.

\subsubsection{Monopoly Principle for Real Task.}

In order to compute this value accurately, we proceed to train the network with images from different scenes.  However, comparing exposures in images from different scenes poses a challenge. Instead of directly contrasting the input and output images, we compare two outputs from the same input, ensuring scene consistency and enabling pixel-to-pixel optimization. Notably, if an over-exposed image is selected as a reference, the resulting image will be brighter; conversely, choosing an under-exposed image as a reference will lead to a darker result. Therefore, by selecting reference images with varying EVs, we can create a bright-dark image pair. We call this ``Monopoly Principle'' to indicate that the output's brightness should be monopoly with the exposure of reference image.

We select two reference images, \(J_1\) and \(J_2\), from a sequence $J$, distinct from $I$, where we sample input images \(I_1\). We assume the EV of \(J_1\) exceeds that of \(J_2\), which means every pixel value of \(J_1\) should be equal or over that of \(J_2\). This condition can be expressed as:
\begin{equation}\label{eq:ev}
\forall (x, y), \quad J_1(x, y) \geq J_2(x, y)  \quad \text{with EV}(J_1) > \text{EV}(J_2).
\end{equation}
Employing \(J_1\) and \(J_2\) as reference images, we apply $d(\cdot, \cdot)$ and $f(\cdot)$ to the same image \(I_1\), resulting in two transformed images, \(I'_{J1}\) and \(I'_{J2}\), respectively. In this case, \(I'_{J1}\) is expected to exhibit a brighter appearance compared to \(I'_{J2}\). This can be described as
\begin{equation}  \label{eq:I'_J1}
    I'_{J1} = f(d(e(I_1), e(J_1)), I_1),\quad I'_{J2} = f(d(e(I_1), e(J_2)), I_1).
\end{equation}
Combine Eq.~(\ref{eq:ev}) and Eq.~(\ref{eq:I'_J1}), we can get
\begin{equation} 
\forall (x, y), \quad I'_{J1}(x, y) \geq I'_{J2}(x, y)  \quad \text{with EV}(J_1) > \text{EV}(J_2).
\end{equation}

\begin{figure}[t]
    \centering
    \begin{subfigure}[b]{0.45\linewidth}
        \includegraphics[width=\linewidth]{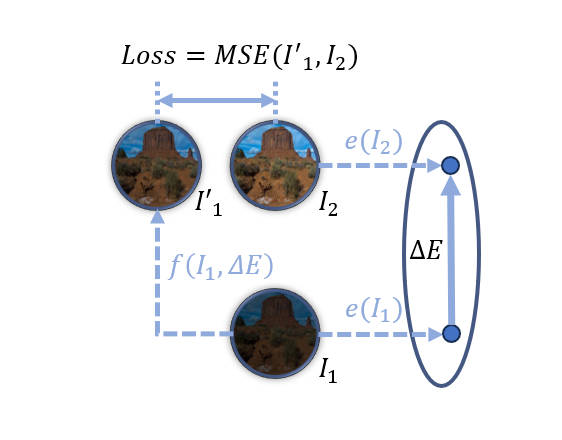}
        \caption{Pretext Task training. Notations are defined in Eq.~(\ref{eq:Delta_E}) and Eq.~(\ref{eq:I'_1})}
        \label{fig:modeling1}
    \end{subfigure}
    \hfill % 用于在子图之间添加空白
    \begin{subfigure}[b]{0.45\linewidth}
        \includegraphics[width=\linewidth]{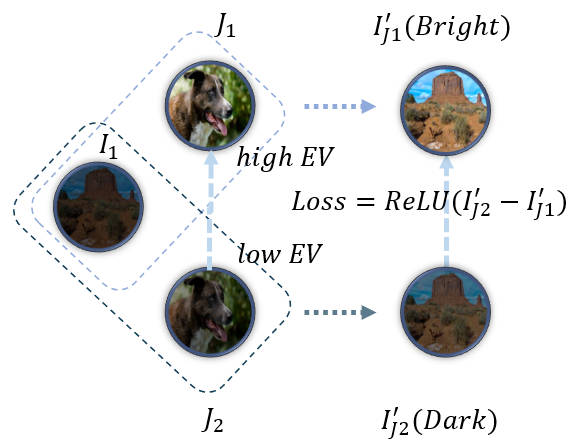}
        \caption{Real Task training.  Notations are defined in Eq.~(\ref{eq:I'_J1})}
        \label{fig:modeling2}
    \end{subfigure}
    \caption{Schematic diagrams of traing UEC.}
    \label{fig:Schematic}
\end{figure}

As demonstrated in Fig.~\ref{fig:modeling2}, this learning criterion enables the network to align exposures across diverse scenes, thereby enhancing the applicability of our method in real-world situations.

% wording of your result has many superfluous words that are inappropriate for a scientific article
\subsubsection{Summarization.} \hu{We design a transformation to adjust exposures, which can be optimized by two principles. Firstly, we select images from a single multi-exposure sequence, establishing a baseline for exposure manipulation. Secondly, we aim to adapt to varied scenes. This involves selecting two reference images from different exposures, but differing from the sequence where we sample the input image. By preserving the relative brightness in the outputs, we facilitate exposure style transformations across different scenes. We concurrently train the network towards using two specific loss functions for each goal.}

\begin{figure}[t]
    \begin{subfigure}[t]{0.24\linewidth}
    \includegraphics[width=\linewidth]{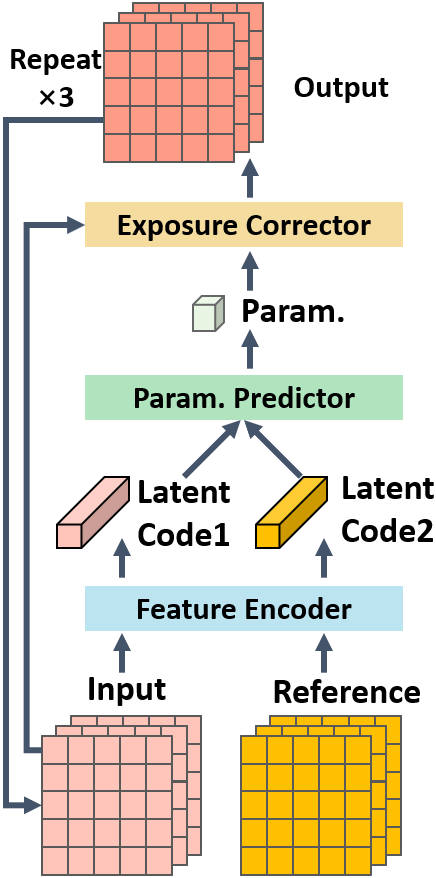}
    \caption{Overview}
    \label{fig:overview}
    \end{subfigure}
    % \vspace{1em}
    \begin{subfigure}[t]{0.24\linewidth}
    \includegraphics[width=\linewidth]{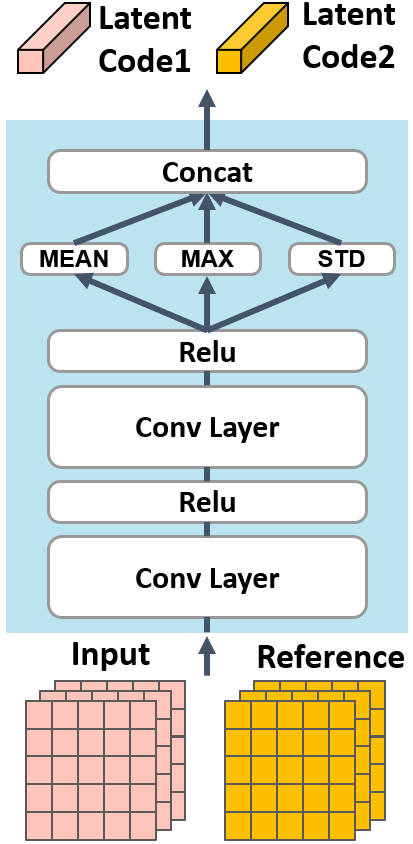}
    \caption{Feature Encoder}
    \label{fig:encoder}
    \end{subfigure}
    % \vspace{1em}
    \begin{subfigure}[t]{0.24\linewidth}
    \includegraphics[width=\linewidth]{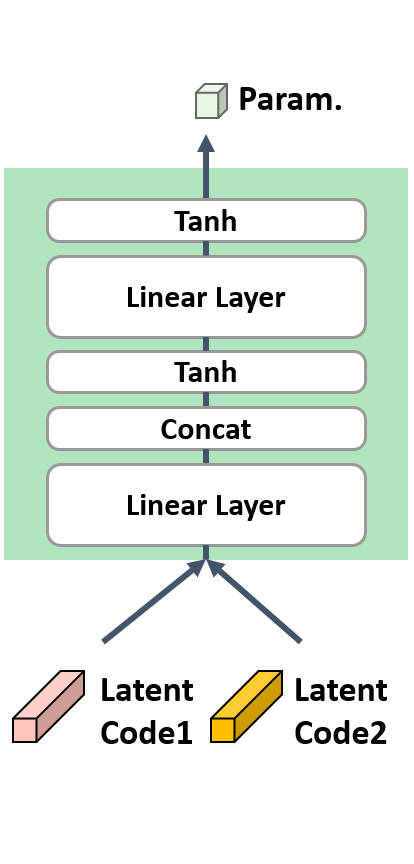}
    \caption{Param. Predictor}
    \label{fig:difference_predictor}
    \end{subfigure}
    % \vspace{1em}
    \begin{subfigure}[t]{0.24\linewidth}
    \includegraphics[width=\linewidth]{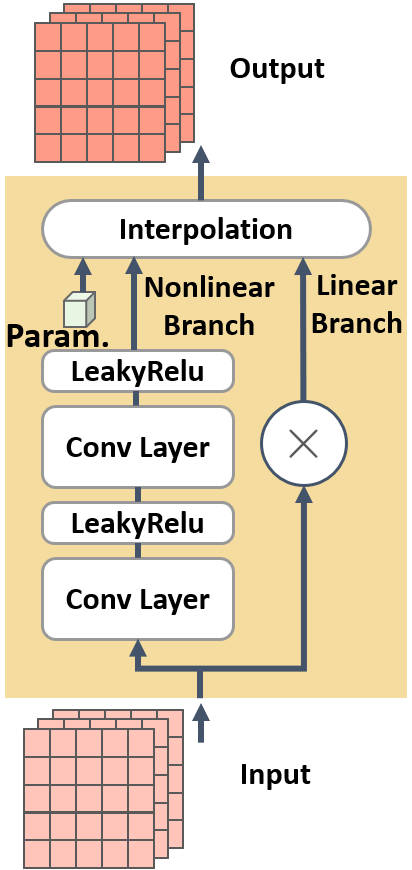}
    \caption{Exp. Corrector}
    \label{fig:corrector}
    \end{subfigure}
    \caption{The architecture of our UEC method.}
    \label{fig:Architecture}
\end{figure}

\subsection{Unsupervised Exposure Correction Architecture} \label{sec:Model Architecture}

We have introduced three crucial functions: (1) $e(\cdot)$ for the extraction of exposure-related features, (2) $d(\cdot, \cdot)$ for the computation of exposure difference between two images in a latent space, and (3) $f(\cdot)$ for the exposure correction of images given the exposure difference. Our UEC network is structured around these functions, as illustrated in the overview shown in Fig.~\ref{fig:overview}. Specifically, our UEC framework incorporates three distinct neural networks: (1) Exposure Feature Encoder, detailed in Fig.~\ref{fig:encoder}; (2) Parameter Predictor based on the difference, detailed in Fig.~\ref{fig:difference_predictor}; and (3) Exposure Corrector, detailed in Fig.~\ref{fig:corrector}. 

Firstly, a decoder comprises two convolutional layers followed by a global pooling layer, producing a 96D feature representation through the combination of maximum, average, and standard deviation across channels. These statistical metrics links to global attributes of images, e.g., contrast, histogram distribution. Secondly, Parameter Predictor evaluates the exposure difference between image pairs, and then computes the parameter $\lambda$ for exposure correction. Thirdly, the exposure correction is modeled as an interpolation of direct scaling and a non-linear adjustment through a network:
\begin{equation}\label{eq:lambda}
I_{out}(x, y) = \lambda \times I_{in}(x, y) + (1 - \lambda) \times h(I_{in}(x, y)),
\end{equation}
where $\lambda$ modulates the transformed between the original input image $I_{in}$ and its corresponding output $I_{out}$. $h(\cdot)$ is the non-linear transformation function implemented by $1 \times 1 $ convolution layers. This iterative process is repeated three times to enhance the correction effectiveness. For more details, please refer to supplementary material.

\subsection{Loss Function}  \label{sec:Loss Function}

We jointly train the Exposure Feature Encoder, Parameter Predictor, and Exposure Corrector in an end-to-end fashion. We employ three types of losses and compute their weighted sum as the final loss.

\subsubsection{Restoration Loss.} Based on the Restoration Supervision discussed in Sec.~\ref{sec:TheoryModeling}, when we sample the input and reference from the same multi-exposure sequence, the synthesized exposure-adjusted image should be identical to the reference image. This is illustrated with $I_1$ and $I_2$ in Fig.~\ref{fig:modeling1}. We utilize the L2 loss for the restoration:

\begin{equation}
    L_{\text{restoration}} = \frac{1}{CHW} \left\lVert I^{\text{out}} - I^{\text{ref}} \right\rVert_2.
\end{equation}
Here, $C$, $H$, and $W$ denote the image's channels, height, and width, respectively. $I^{\text{out}}$ is the neural network output, and $I^{\text{ref}}$ the reference image.

\subsubsection{Monopoly Loss.} Based on the Monopoly Principle discussed in Sec.~\ref{sec:TheoryModeling}, when we sample the two references from the same multi-exposure sequence, which is different from the input's sampling source, the relative brightness relation should also be transferred to the output image pair. This is shown with $I'_{J1}$ and $I'_{J2}$ in Fig.~\ref{fig:modeling2}. We utilize the ReLU function for the monopoly loss:

\begin{equation}
    L_{\text{monopoly}} = \frac{1}{CHW} \text{ReLU} \left( I^{\text{out2}} - I^{\text{out1}} \right) \quad \text{with EV}(I^{\text{ref1}}) > \text{EV}(I^{\text{ref2}}).
\end{equation}
In this equation, $I^{\text{out1}}$ and $I^{\text{out2}}$ represent the neural network outputs corresponding to the references $I^{\text{ref1}}$ and $I^{\text{ref2}}$, respectively. It is important to note that the EV of $I^{\text{ref1}}$ is greater than that of $I^{\text{ref2}}$. The ReLU function ensures that the loss calculation emphasizes the positive differences, aligning with the expected brightness relation between the two outputs.

\subsubsection{Semantic-preserving Loss.} For the preservation of semantics, we use total variation loss\cite{aly2005image}, a proven regularization technique to maintain spatial coherence and semantic integrity. This approach smooths transitions between adjacent pixels:

\begin{equation}
    L_{\text{semantic}} = \frac{1}{CHW} \left\lVert \nabla I^{\text{out}} \right\rVert_2.
\end{equation}
Here, $\nabla(\cdot)$ denotes the gradient operator, computing the image's spatial derivative. 

% \subsubsection{Style Loss.}
% In addition to semantic consistency, color congruity is crucial for producing visually appealing images. We introduce a color loss component \cite{liu2020color, wang2019underexposed}, which quantifies angular discrepancies between RGB colors, considered as 3D vectors:

% \begin{equation}
%     L_{\text{style}} = 1 - \frac{1}{HW} \angle(I^{\text{out}}, I^{\text{ref}}).
% \end{equation}

% where $\angle(\cdot)$ calculates the average cosine of angular differences across pixels. This careful integration of various loss terms aligns with our approach's overarching goals. 

\subsubsection{Summarization.} The final loss function is expressed as:

\begin{equation} \label{eq:final}
    L = \alpha_1 \cdot L_{\text{restoration}} + \alpha_2 \cdot L_{\text{monopoly}} + \alpha_3 \cdot L_{\text{semantic}},
\end{equation}
where $\alpha_1$, $\alpha_2$, and $\alpha_3$ are the balancing weights. Since our framework (Exposure Feature Encoder, Parameter Predictor, Exposure Corrector) works in an end-to-end way, the total loss function, Eq.~(\ref{eq:final}), is added on the end of exposure corrector to update the weights of all the modules during training.

\subsection{Testing Details}  \label{sec:test}

We have outlined the training procedure for our UEC model, which employs reference images of diverse quality to address the full spectrum of exposure adjustments. For the testing phase, it is crucial to identify the optimal value for the final calibration. The testing inference largely replicates the training process as depicted in Fig.~\ref{fig:overview}, with the primary distinction being the use of a single reference image for all testing inputs. Specifically, we hard code the exposure features derived from this image across all test cases.

The efficiency of this method is underscored by its minimal data requirements. Mastering exposure adjustments is complex, but identifying an image's optimal exposure level is considerably more straightforward. Our approach emphasizes radiometry to ensure consistency, which differs from colorimetry, thereby minimizing the data needed to such an extent that a single ground truth image is sufficient. After training, the UEC model is adept at generating a sequence of images with varying exposures for any given input. The goal during testing is to select the most suitable image from this sequence. To enhance efficiency, our method bypasses the need to generate and then choose from multiple exposures, instead directly yielding the image with the best exposure.

\section{Our Radiometry Correction Dataset} \label{sec:Our Dataset}

In Sec.~\ref{sec:intro}, we have explored the benefits of disentanglement in radiometry adjustment for enhancing the generalizability of exposure correction. For this purpose, a specialized dataset is essential. Hence, we developed Radiometry Correction Dataset using MIT-Adobe FiveK Dataset \cite{bychkovsky2011learning}, which includes 5,000 RAW photographs and their expertly modified sRGB images. These adjustments were performed by five specialists who worked directly with the RAW images. Drawing from their expertise, we adopted a reverse engineering approach to generate ill-exposed images. This process entails adjusting the exposure of the reference images to generate synthetic versions that are either underexposed or overexposed, while while freezing other post-processing ISP procedures to keep the individual style constant. Following the established convention, we selected the versions edited by ExpertC as our reference images. Based on this, we adjusted the exposures relative to the original images, spanning -2EV, -1EV, 0EV, +1EV, +2EV, and +3EV. Considering that input images are often underexposed, this range ensures a balanced exposure spectrum in relation to the reference images.

\begin{figure*}[t]
    \centering
    \begin{overpic}[width=0.12\linewidth, height=0.07\linewidth]{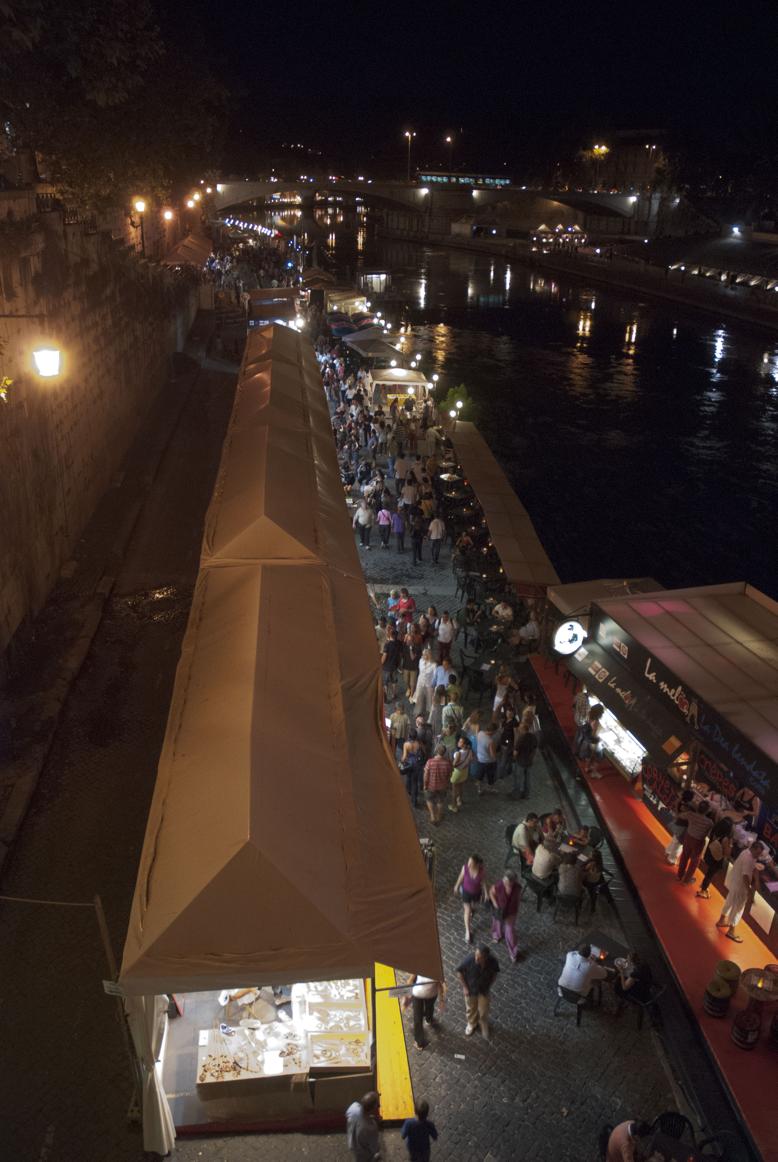}
        \put(2,45){\textcolor[rgb]{1,1,1}{\tiny -1.5EV}}
    \end{overpic}
    \begin{overpic}[width=0.12\linewidth, height=0.07\linewidth]{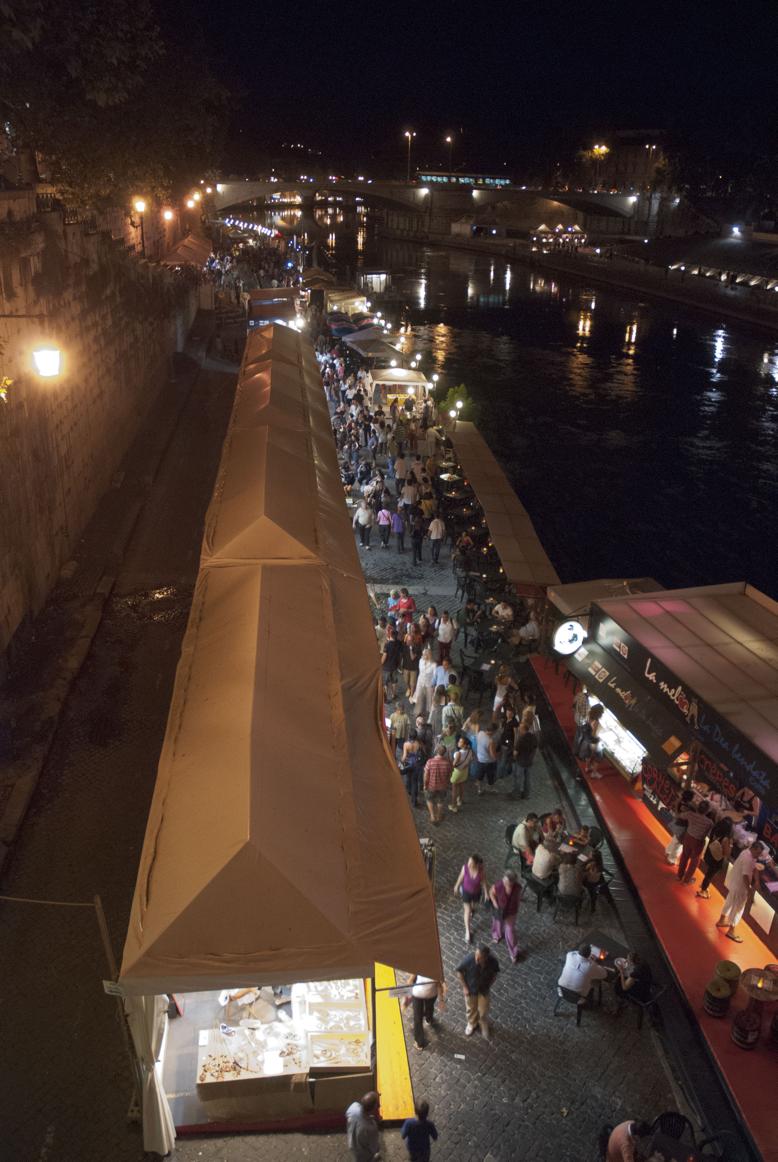}
        \put(2,45){\textcolor[rgb]{1,1,1}{\tiny -1.0EV}}
    \end{overpic}
    \begin{overpic}[width=0.12\linewidth, height=0.07\linewidth]{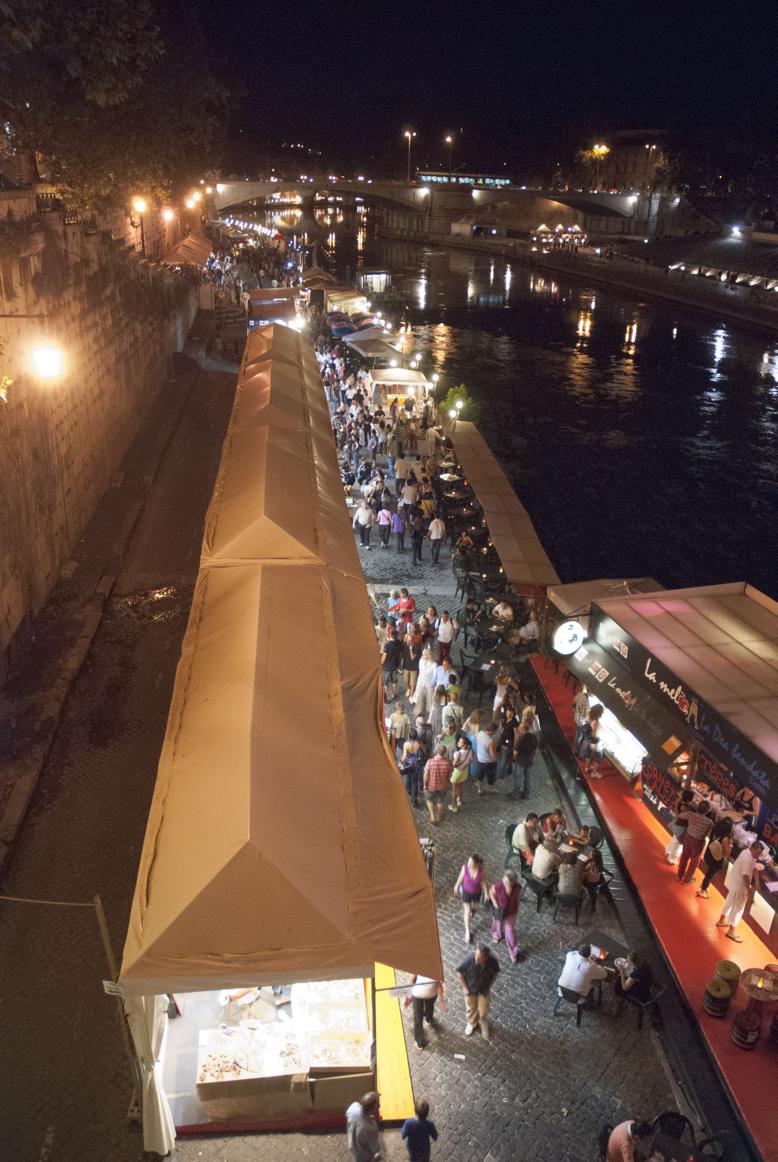}
        \put(2,45){\textcolor[rgb]{1,1,1}{\tiny 0EV}}
    \end{overpic}
    \begin{overpic}[width=0.12\linewidth, height=0.07\linewidth]{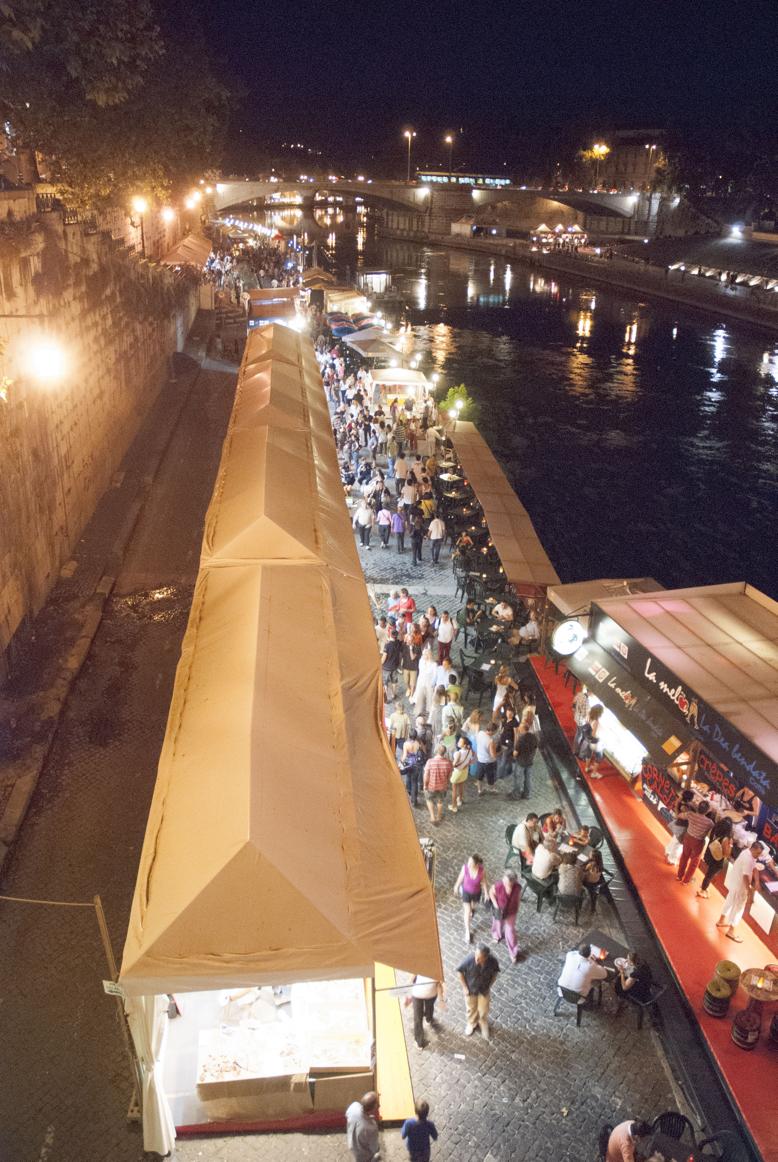}
        \put(2,45){\textcolor[rgb]{1,1,1}{\tiny +1.0EV}}
    \end{overpic}
    \begin{overpic}[width=0.12\linewidth, height=0.07\linewidth]{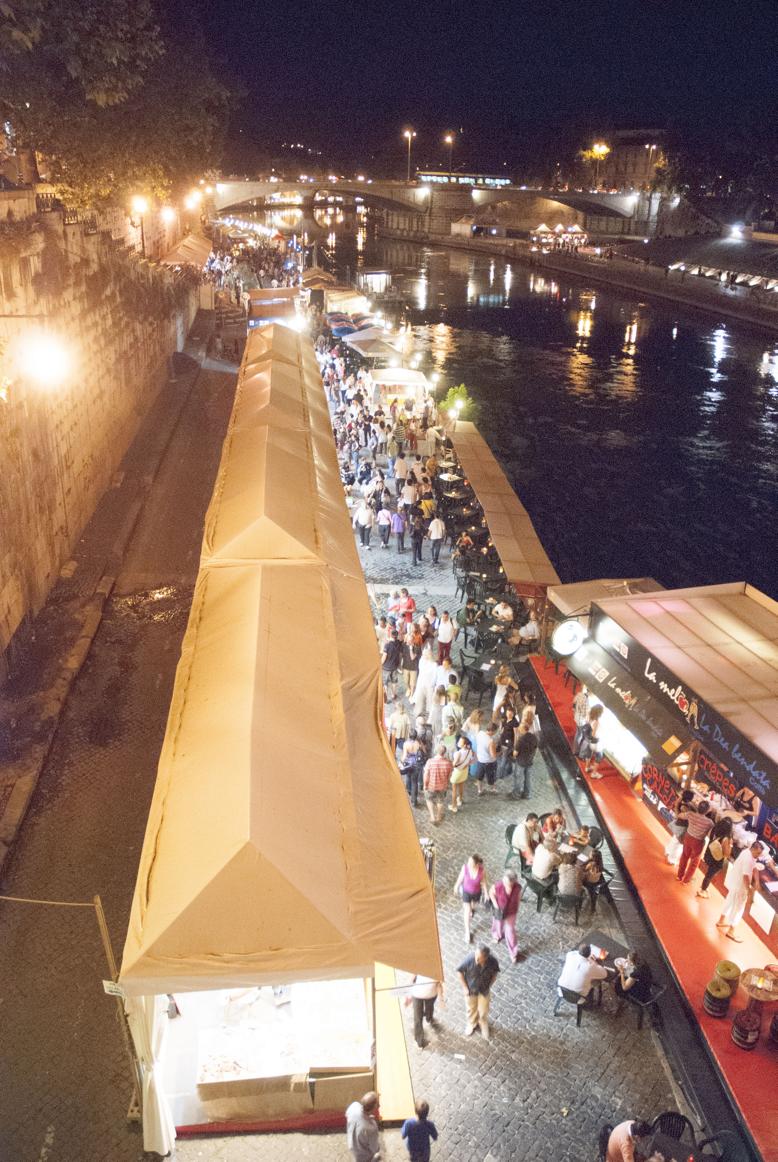}
        \put(2,45){\textcolor[rgb]{1,1,1}{\tiny +1.5EV}}
    \end{overpic}
    \begin{overpic}[width=0.12\linewidth, height=0.07\linewidth]{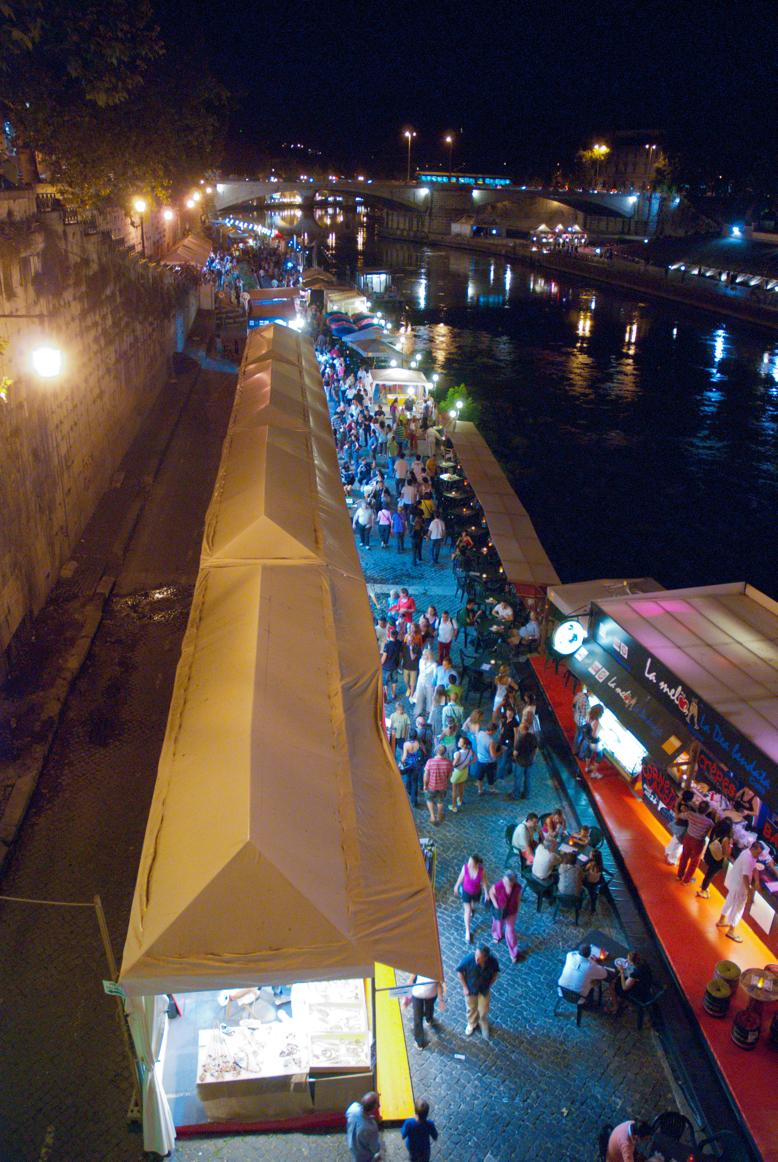}
        \put(2,45){\textcolor[rgb]{1,1,1}{\tiny GT}}
    \end{overpic}
    \\
    \centering
    \scriptsize (a) Afifi et al's MSEC Dataset \cite{Afifi2021}
    \\
    \begin{overpic}[width=0.12\linewidth, height=0.07\linewidth]{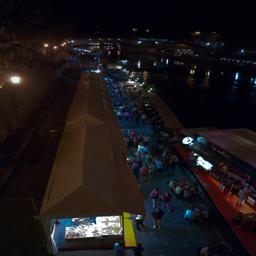}
        \put(2,45){\textcolor[rgb]{1,1,1}{\tiny -2EV}}
    \end{overpic}
    \begin{overpic}[width=0.12\linewidth, height=0.07\linewidth]{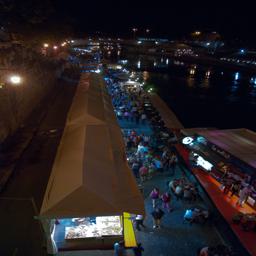}
        \put(2,45){\textcolor[rgb]{1,1,1}{\tiny -1EV}}
    \end{overpic}
    \begin{overpic}[width=0.12\linewidth, height=0.07\linewidth]{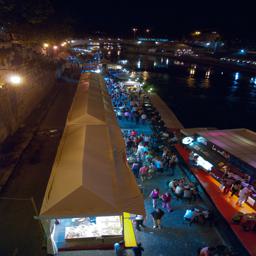}
        \put(2,45){\textcolor[rgb]{1,1,1}{\tiny 0EV}}
    \end{overpic}
    \begin{overpic}[width=0.12\linewidth, height=0.07\linewidth]{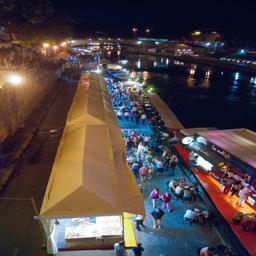}
        \put(2,45){\textcolor[rgb]{1,1,1}{\tiny +1EV}}
    \end{overpic}
    \begin{overpic}[width=0.12\linewidth, height=0.07\linewidth]{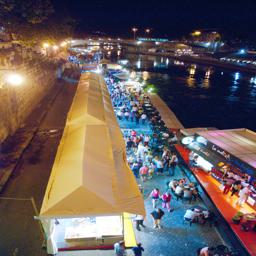}
        \put(2,45){\textcolor[rgb]{1,1,1}{\tiny +2EV}}
    \end{overpic}
    \begin{overpic}[width=0.12\linewidth, height=0.07\linewidth]{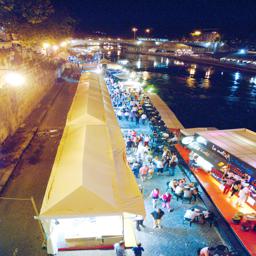}
        \put(2,45){\textcolor[rgb]{1,1,1}{\tiny +3EV}}
    \end{overpic}
    \begin{overpic}[width=0.12\linewidth, height=0.07\linewidth]{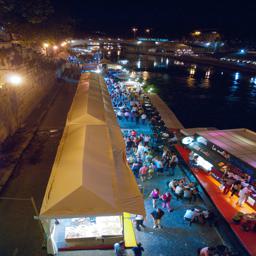}
        \put(2,45){\textcolor[rgb]{1,1,1}{\tiny GT}}
    \end{overpic}
    \\
    \centering
    \scriptsize (b) Our Radiometry Correction Dataset
    \\
    \caption{
        Visualization of Datasets: (a) Afifi et al's \cite{Afifi2021} vs. (b) Ours.
    }
    \label{fig:dataset_cmp}
\end{figure*}

Fig.~\ref{fig:dataset_cmp} provides a comparison between Afifi's MSEC Dataset \cite{Afifi2021} and our Radiometry Correction Dataset. Evidently, in this example, the ground exhibits a blue hue as ExpertC's individual style. For additional comparisons, please consult the supplementary material.

\section{Experiments}

\subsection{Experiment Settings}

\subsubsection{Datasets.} The prevailing benchmark primarily utilizes the MSEC Dataset from the study by Afifi et al. \cite{Afifi2021}. To mitigate bias stemming from individual styles on colorimetry, we introduce our Radiometry Correction Dataset. Both datasets are employed for training and testing various methods. To assess the models' generalizability, we conduct training on the Exposure Dataset and perform testing on LOL dataset \cite{cai2018learning}, which is specifically designed for low-light image enhancement.

\subsubsection{Implementation Details.} The weights in Eq.~(\ref{eq:final}) are empirically set $\alpha_1 = \alpha_2 = 1$ and $\alpha_3 = 0.1$. We use only one well-exposed reference image during testing.  We select the second image from left to right in Fig.~\ref{fig:ref_imgs} as our reference in our evaluations. More details are described in our supplementary material. %We set $\beta_1 = 0.9, \beta_2 = 0.999$. 

\subsubsection{Evaluation Metrics.} In Sec.~\ref{sec:intro}, we highlighted the importance of evaluating exposure correction in terms of both overall aesthetics and low-level features. To assess overall aesthetics, we compare the output to the ground truth images using metrics such as PSNR and SSIM \cite{sara2019image}. For low-level feature evaluation, we analyze the output through edge detection and make comparisons using PSNR and F1 scores, applying a threshold based on median pixel values.

%\subsubsection{Selection of Reference Images.} If we do not specify, 

\subsection{Exposure Correction Results}

\subsubsection{Results on MSEC Dataset.} 

We compare our results with the ground truth, as presented in Tab.~\ref{tab:Comparison_ECD}. Even though our Unsupervised Exposure Correction (UEC) model has not been trained on manually calibrated data and our adjustments are confined to radiometry, the proficiency of our UEC model remains competitive when compared to supervised SOTA models like ECM \cite{Eyiokur2022} and Afifi et al. \cite{Afifi2021}.

\begin{figure}[t]
    % \parbox[t][0.02\linewidth][c]{0.45\linewidth}{\scriptsize \centering (a) Input Multi-exposure Images}
    % \parbox[t][0.02\linewidth][c]{0.45\linewidth}{\scriptsize \centering (b) Ground Truth with Stylistic Biases}
    \begin{subfigure}[t]{0.48\linewidth}
    % \centering
    \begin{overpic}[width=0.18\textwidth]{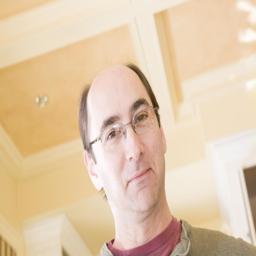}
        \put(2,85){\textcolor[rgb]{1,1,1}{\tiny +1.5EV}}
    \end{overpic}
    \begin{overpic}[width=0.18\textwidth]{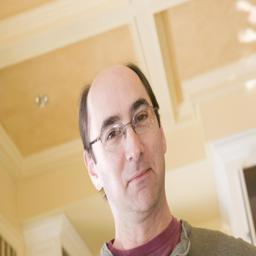}
        \put(2,85){\textcolor[rgb]{1,1,1}{\tiny +1.0EV}}
    \end{overpic}
    \begin{overpic}[width=0.18\textwidth]{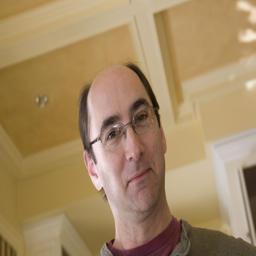}
        \put(2,85){\textcolor[rgb]{1,1,1}{\tiny 0EV}}
    \end{overpic}
    \begin{overpic}[width=0.18\textwidth]{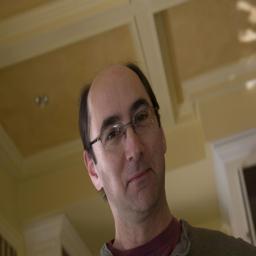}
        \put(2,85){\textcolor[rgb]{1,1,1}{\tiny -1.0EV}}
    \end{overpic}
    \begin{overpic}[width=0.18\textwidth]{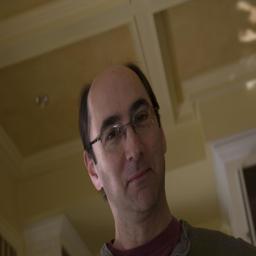}
        \put(2,85){\textcolor[rgb]{1,1,1}{\tiny -1.5EV}}
    \end{overpic}
    \caption{Input Multi-exposure Sequence}
    \label{fig:artifact_input2}
    \end{subfigure}
    \begin{subfigure}[t]{0.48\linewidth}
    \begin{overpic}[width=0.18\textwidth]{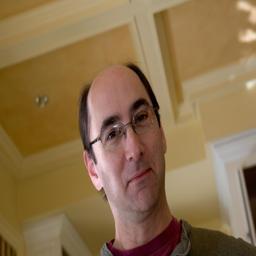}
        \put(2,85){\textcolor[rgb]{1,1,1}{\tiny ExpertA}}
    \end{overpic}
    \begin{overpic}[width=0.18\textwidth]{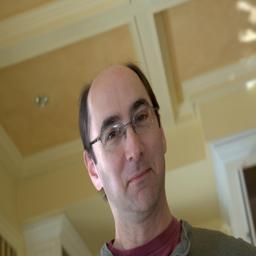}
        \put(2,85){\textcolor[rgb]{1,1,1}{\tiny ExpertB}}
    \end{overpic}
    \begin{overpic}[width=0.18\textwidth]{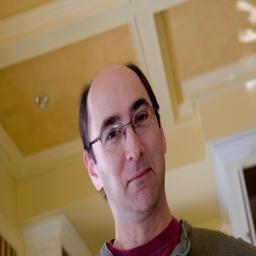}
        \put(2,85){\textcolor[rgb]{1,1,1}{\tiny ExpertC}}
    \end{overpic}
    \begin{overpic}[width=0.18\textwidth]{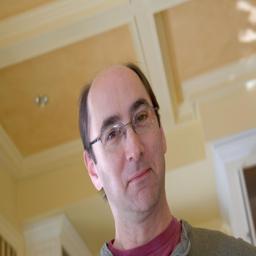}
        \put(2,85){\textcolor[rgb]{1,1,1}{\tiny ExpertD}}
    \end{overpic}
    \begin{overpic}[width=0.18\textwidth]{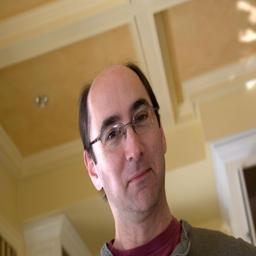}
        \put(2,85){\textcolor[rgb]{1,1,1}{\tiny ExpertE}}
    \end{overpic}
    \caption{Ground Truths with Stylistic Biases}
    \label{fig:artifact_biasGT2}
    \end{subfigure}
    \\
    % \parbox[t][0.02\linewidth][c]{0.45\linewidth}{\scriptsize \centering (c) ECM \cite{Eyiokur2022}(Sup.)}
    % \parbox[t][0.02\linewidth][c]{0.45\linewidth}{\scriptsize \centering (d) Ours(Unsup.) }
    \begin{subfigure}[t]{0.48\linewidth}
    \begin{overpic}[width=0.18\textwidth]{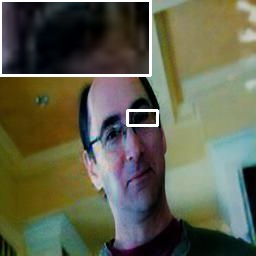}
    \end{overpic}
    \begin{overpic}[width=0.18\textwidth]{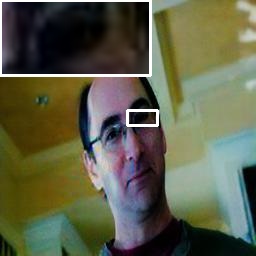}
    \end{overpic}
    \begin{overpic}[width=0.18\textwidth]{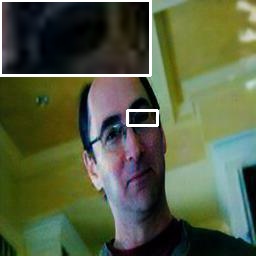}
    \end{overpic}
    \begin{overpic}[width=0.18\textwidth]{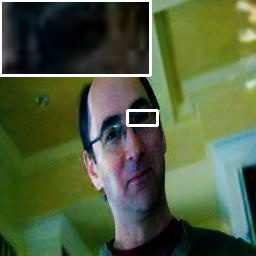}
    \end{overpic}
    \begin{overpic}[width=0.18\textwidth]{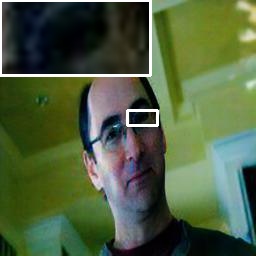}
    \end{overpic}
    \caption{ECM \cite{Eyiokur2022}(Sup.)}
    \label{fig:artifact_ecm2}
    \end{subfigure}
    \begin{subfigure}[t]{0.48\linewidth}
    \begin{overpic}[width=0.18\textwidth]{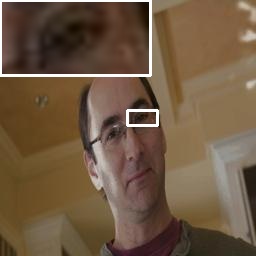}
    \end{overpic}
    \begin{overpic}[width=0.18\textwidth]{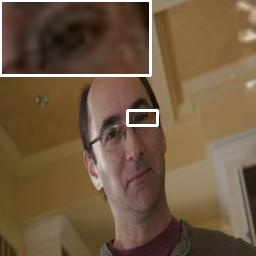}
    \end{overpic}
    \begin{overpic}[width=0.18\textwidth]{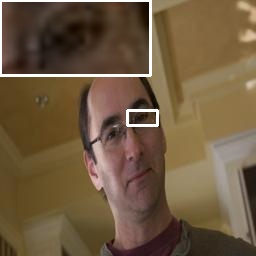}
    \end{overpic}
    \begin{overpic}[width=0.18\textwidth]{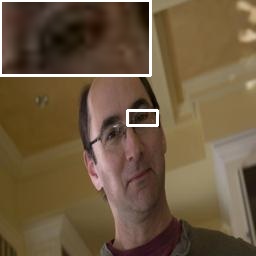}
    \end{overpic}
    \begin{overpic}[width=0.18\textwidth]{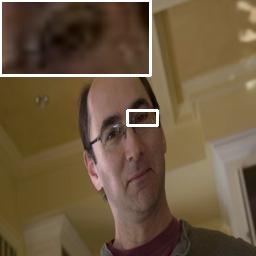}
    \end{overpic}
    \caption{Ours(Unsup.)}
    \label{fig:artifact_ours2}
    \end{subfigure}
    \caption{Visual Comparison: ECM \cite{Eyiokur2022} vs. Our UEC method.}
    \label{fig:artifact2}
\end{figure}

\begin{figure}[t]
    \centering
    \parbox[t][1em][c]{0.0675\textwidth}{\tiny \centering Input}
    \parbox[t][1em][c]{0.0675\textwidth}{\tiny \centering DPED }
    \parbox[t][1em][c]{0.0675\textwidth}{\tiny \centering Afifi et al. \cite{Afifi2021}}
    \parbox[t][1em][c]{0.0675\textwidth}{\tiny \centering ECM \cite{Eyiokur2022}}
    \parbox[t][1em][c]{0.0675\textwidth}{\tiny \centering Ours}
    \parbox[t][1em][c]{0.0675\textwidth}{\tiny \centering GT}
    \smallskip
    \parbox[t][1em][c]{0.0675\textwidth}{\tiny \centering Input}
    \parbox[t][1em][c]{0.0675\textwidth}{\tiny \centering DPED }
    \parbox[t][1em][c]{0.0675\textwidth}{\tiny \centering Afifi et al. \cite{Afifi2021}}
    \parbox[t][1em][c]{0.0675\textwidth}{\tiny \centering ECM \cite{Eyiokur2022}}
    \parbox[t][1em][c]{0.0675\textwidth}{\tiny \centering Ours}
    \parbox[t][1em][c]{0.0675\textwidth}{\tiny \centering GT}
    \centering
    \includegraphics[width=0.45\textwidth]{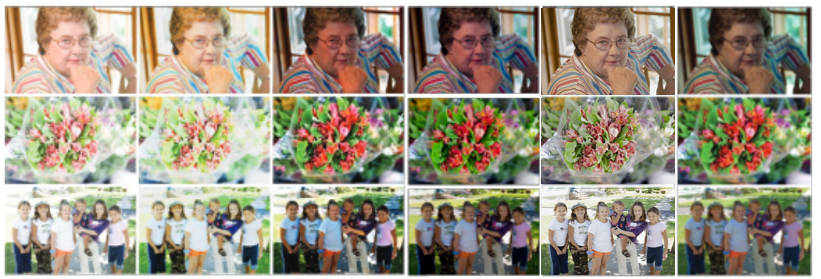}
    \smallskip
    \includegraphics[width=0.45\textwidth]{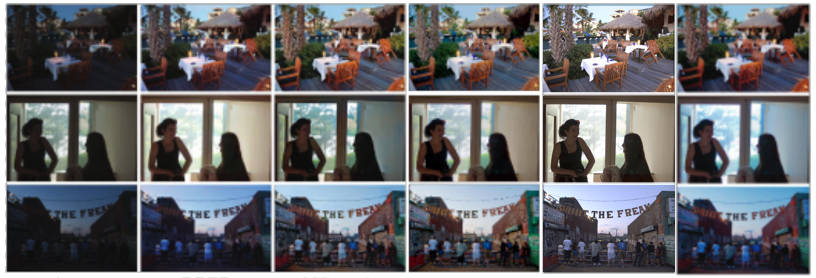}
    \\
    \caption{Results on MSEC Dataset \cite{Afifi2021}. We take images from \cite{Eyiokur2022} and compare with our model.}
    \label{fig:cmp_many_methods}
\end{figure}

In Fig.~\ref{fig:cmp_many_methods}, we present a comparison of test images using various methods, including our approach, ground truth images, and results from supervised models. Our method demonstrates competitive performance in radiometry correction, matching the SOTA supervised techniques. Furthermore, in Fig.~\ref{fig:artifact} and Fig.~\ref{fig:artifact2}, we compare our UEC method with ECM \cite{Eyiokur2022}. In Fig.~\ref{fig:artifact}, it is important to note that the ground truth colorimetry, as determined by five experts, exhibits variability, indicating that learning from human-adjusted images may introduce biases reflective of their individual styles. In Fig.~\ref{fig:artifact2}, the ECM’s results\cite{Eyiokur2022} (Fig.~\ref{fig:artifact_ecm2}) manifest a color cast in the background, where the background turns green. Our method addresses this issue by applying radiometry correction. Furthermore, to highlight the superior detail resolution achieved by our UEC method, we have magnified a segment of the image located in the upper left corner. Further comparisons are elaborated in the supplementary material.

\begin{table}[t]  
\centering  
\begin{subtable}{.45\textwidth}  
  \centering  
  \begin{tabular}{lll}  
    \toprule 
    Method & PSNR & SSIM \\
    \midrule
    HDRCNN w/PS \cite{dayley2010photoshop} & 17.032 & 0.687 \\
    DPED (iPhone) \cite{ignatov2017dslr} & 16.274 & 0.629 \\
    DPED (BlackBerry) \cite{ignatov2017dslr} & 17.890 & 0.671 \\
    DPE(HDR) \cite{chen2018deep} & 16.206 & 0.623 \\
    DPE(S-FiveK) \cite{chen2018deep} & 17.510 & 0.677 \\
    % RetinexNet* \cite{wei2018deep} & 11.633 & 0.607 \\
    % DeepUPE*\cite{chen2018deep} & 14.247 & 0.640 \\
    Zero-DCE \cite{guo2020zero} & 12.597 & 0.549 \\
    Afifi et al. \cite{Afifi2021} & 19.483 & 0.739 \\ 
    ECM \cite{Eyiokur2022} & \textbf{20.874} & \textbf{0.877} \\
    Ours & 18.756 & 0.812 \\
    \bottomrule
  \end{tabular}
  \caption{Testing Result} 
  % \caption{Performance on MSEC Dataset} 
  \label{tab:Comparison_ECD}
\end{subtable}%  
\hfill % 使用\hfill来最大化两个子表之间的水平间距  
\begin{subtable}{.45\textwidth}
  \centering
  \begin{tabular}{lccc}
    \toprule
    Method & PSNR & SSIM \\
    \midrule
    Afifi et al.(E) \cite{Afifi2021} & 14.268 & 0.638 \\
    ECM(E) \cite{Eyiokur2022} & 15.439 & 0.650 \\
    ECM(R) \cite{Eyiokur2022} & 17.537 & 0.725 \\
    Ours(E) & \textbf{18.571} & \textbf{0.728} \\
    \bottomrule
  \end{tabular}
  \caption{generalizability Results. ``E'' denotes pretraining on MSEC Dataset \cite{Afifi2021}; ``R'' signifies pretraining on our Radiometry Correction Dataset. We use LOL dataset \cite{wei2018deep} for evaluation. }
  \label{tab:Generalization}
\end{subtable} 
\caption{Quantitative comparison of performance and generalizability.}  
\label{tab:cmp_ecd}  
\end{table}

\subsubsection{Results on Generalizability.} 

% As detailed in Sec.\ref{sec:intro}, the supervision in the Exposure Dataset is influenced by the individual styles of various experts, posing challenges to the generalization capacity of models. 

To assess the models' generalizability, we utilized pretrained models and evaluated their performance on LOL dataset \cite{wei2018deep}. LOL dataset comprises 485 pairs of training images, with each pair featuring one low-light image and one corresponding normal-light image. For our assessment, we exclusively employed the training image pairs from LOL dataset to serve as our evaluation set.

\begin{figure}[t]
    \centering
    \parbox[t][0.05\textwidth][c]{0.10\textwidth}{\scriptsize \centering Input}
    \parbox[t][0.05\textwidth][c]{0.10\textwidth}{\scriptsize \centering ECM \cite{Eyiokur2022}\\(Sup.)}
    \parbox[t][0.05\textwidth][c]{0.10\textwidth}{\scriptsize \centering Ours\\(Unsup.)}
    \parbox[t][0.05\textwidth][c]{0.10\textwidth}{\scriptsize \centering GT}
    \parbox[t][0.05\textwidth][c]{0.10\textwidth}{\scriptsize \centering Input}
    \parbox[t][0.05\textwidth][c]{0.10\textwidth}{\scriptsize \centering ECM \cite{Eyiokur2022}\\(Sup.)}
    \parbox[t][0.05\textwidth][c]{0.10\textwidth}{\scriptsize \centering Ours\\(Unsup.)}
    \parbox[t][0.05\textwidth][c]{0.10\textwidth}{\scriptsize \centering GT}
    \\
    \includegraphics[width=0.10\textwidth]{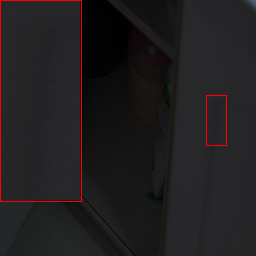}
    \includegraphics[width=0.10\textwidth]{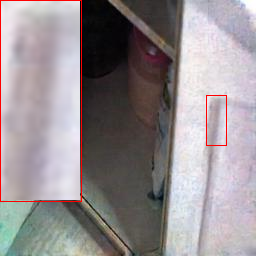}
    \includegraphics[width=0.10\textwidth]{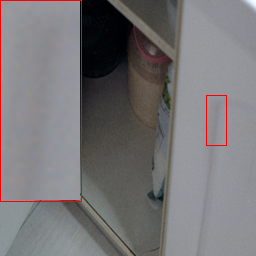}
    \includegraphics[width=0.10\textwidth]{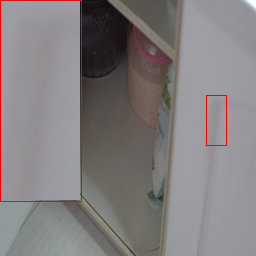}
    \includegraphics[width=0.10\textwidth]{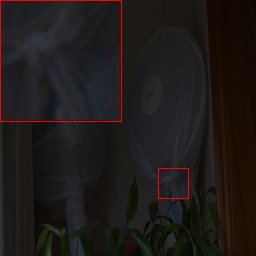}
    \includegraphics[width=0.10\textwidth]{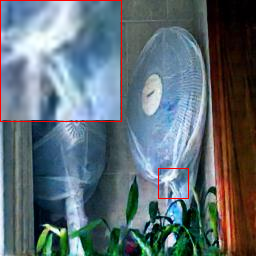}
    \includegraphics[width=0.10\textwidth]{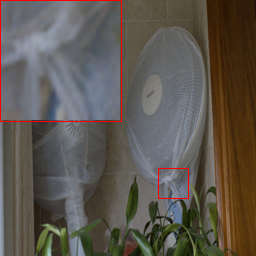}
    \includegraphics[width=0.10\textwidth]{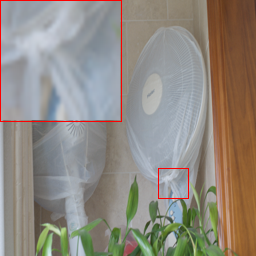}
    \\
    \includegraphics[width=0.10\textwidth]{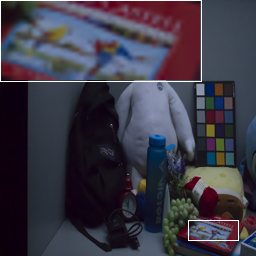}
    \includegraphics[width=0.10\textwidth]{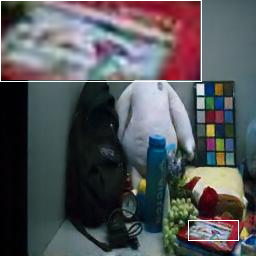}
    \includegraphics[width=0.10\textwidth]{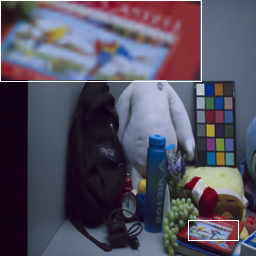}
    \includegraphics[width=0.10\textwidth]{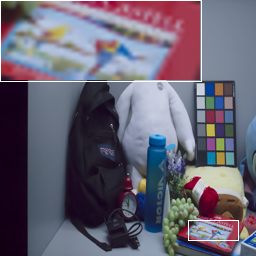}
    \includegraphics[width=0.10\textwidth]{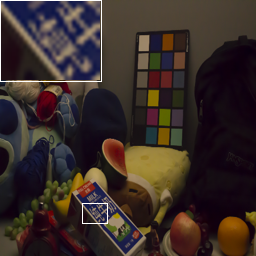}
    \includegraphics[width=0.10\textwidth]{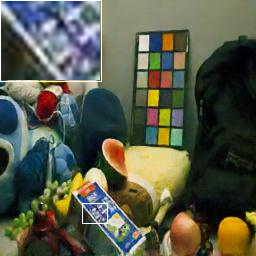}
    \includegraphics[width=0.10\textwidth]{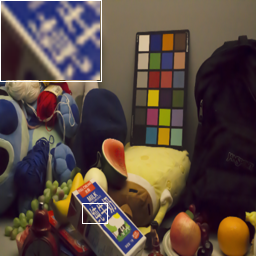}
    \includegraphics[width=0.10\textwidth]{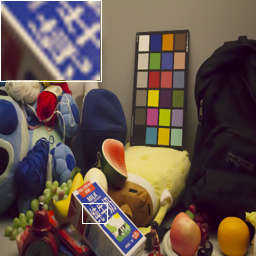}
    \caption{Generalizability performance by ECM \cite{Eyiokur2022} and our UEC. We pretrain models on MSEC Dataset \cite{Afifi2021} and assess their performance on LOL Dataset \cite{wei2018deep}.}

    \label{fig:Generalization}
\end{figure}

For evaluation, we selected one ground truth image from MSEC Dataset to serve as our reference image. As illustrated in Tab.~\ref{tab:Generalization}, while the performance of supervised models significantly decreases, our results demonstrate stability and outperform ECM \cite{Eyiokur2022}. It is noteworthy that ECM \cite{Eyiokur2022}, when trained on our Radiometry Correction Dataset, displayed superior generalizability compared to its performance on MSEC Dataset.
The visualization in Fig.~\ref{fig:Generalization} reveals that our approach ensures better color consistency without stylistic biases. These results underscores the robustness of radiometry correction.

\subsubsection{Results on Radiometry Correction Dataset.} To solely assess radiometric performance, we trained and evaluated exposure correction methods using our Radiometry Correction Dataset. We benchmarked our UEC method against the SOTA supervised model, ECM \cite{Eyiokur2022}. Both methods demonstrated similar performance in terms of PSNR; however, our UEC method significantly outperformed in terms of SSIM. The results are detailed in Tab.~\ref{tab:Comparison_Exposures}. To ensure an fair comparison, we utilized the official implementations and adhered to the default hyperparameter configurations.

\begin{table}[t]  
\centering  
\begin{subtable}{.45\textwidth}  
  \centering  
    \begin{tabular}{cccccc}
    \toprule
     & \multicolumn{2}{c}{ECM \cite{Eyiokur2022}} & \multicolumn{2}{c}{Ours} \\
     EV & PSNR & SSIM & PSNR & SSIM \\
    \midrule
    -2 & \textbf{20.122} & 0.718 & 19.475 & \textbf{0.812} \\
    -1 & \textbf{20.301} & 0.735 & 20.144 & \textbf{0.847} \\
    0 & 20.546 & 0.751 & \textbf{20.968} & \textbf{0.884} \\
    +1 & 20.667 & 0.758 & \textbf{21.614} & \textbf{0.907} \\
    +2 & 20.588 & 0.758 & \textbf{21.187} & \textbf{0.897} \\
    +3 & \textbf{20.445} & 0.743 & 19.901 & \textbf{0.862} \\
    Avg. & 20.445 & 0.744 & \textbf{20.548} & \textbf{0.868} \\
    \bottomrule
    \end{tabular}
    \caption{Results on Radiometry Correction Dataset}
    \label{tab:Comparison_Exposures}
\end{subtable}%  
\hfill % 使用\hfill来最大化两个子表之间的水平间距  
\begin{subtable}{.45\textwidth}
  \centering
    \begin{tabular}{cccccc}
    \toprule
     & \multicolumn{2}{c}{ECM \cite{Eyiokur2022}} & \multicolumn{2}{c}{Ours} \\
     EV & PSNR & F1-Score & PSNR & F1-Score \\
    \midrule
    -2 & 15.510 & 0.912 & \textbf{20.059} & \textbf{0.959} \\
    -1 & 15.983 & 0.918 & \textbf{21.616} & \textbf{0.966} \\
    0 & 16.449 & 0.923 & \textbf{23.746} & \textbf{0.974} \\
    +1 & 16.730 & 0.925 & \textbf{25.550} & \textbf{0.978} \\
    +2 & 16.758 & 0.926 & \textbf{24.090} & \textbf{0.975} \\
    +3 & 16.443 & 0.925 & \textbf{20.929} & \textbf{0.964} \\
    Avg. & 16.312 & 0.922 & \textbf{22.665} & \textbf{0.969} \\
    \bottomrule
    \end{tabular}
    \caption{Edge detection results on Radiometry Correction Dataset.}
    \label{tab:result_edgedet}
\end{subtable} 
\caption{Quantitative comparison of performance and generalizability.}  
\label{tab:cmp_rcd}  
\end{table}

% \begin{table}
% \centering
% \begin{tabular}{cccccc}
% \toprule
%  & \multicolumn{2}{c}{ECM \cite{Eyiokur2022}} & \multicolumn{2}{c}{Ours} \\
%  EV & PSNR & SSIM & PSNR & SSIM \\
% \midrule
% -2 & 20.122 & 0.718 & 19.475 & 0.812 \\
% -1 & 20.301 & 0.735 & 20.144 & 0.847 \\
% 0 & 20.546 & 0.751 & 20.968 & 0.884 \\
% +1 & 20.667 & 0.758 & 21.614 & 0.907 \\
% +2 & 20.588 & 0.758 & 21.187 & 0.897 \\
% +3 & 20.445 & 0.743 & 19.901 & 0.862 \\
% Avg. & 20.445 & 0.744 & 20.548 & 0.868 \\
% \bottomrule
% \end{tabular}
% \caption{Results on Radiometry Correction Dataset}
% \label{tab:Comparison_Exposures}
% \end{table}
% \subsection{Downstream Task Evaluation}

% \begin{table}
% \centering
% \begin{tabular}{cccccc}
% \toprule
%  & \multicolumn{2}{c}{ECM \cite{Eyiokur2022}} & \multicolumn{2}{c}{Ours} \\
%  EV & PSNR & F1-Score & PSNR & F1-Score \\
% \midrule
% -2 & 15.510 & 0.912 & 20.059 & 0.959 \\
% -1 & 15.983 & 0.918 & 21.616 & 0.966 \\
% 0 & 16.449 & 0.923 & 23.746 & 0.974 \\
% +1 & 16.730 & 0.925 & 25.550 & 0.978 \\
% +2 & 16.758 & 0.926 & 24.090 & 0.975 \\
% +3 & 16.443 & 0.925 & 20.929 & 0.964 \\
% \textbf{Avg.} & \textbf{16.312} & \textbf{0.922} & \textbf{22.665} & \textbf{0.969} \\
% \bottomrule
% \end{tabular}
% \caption{Edge detection results on Radiometry Correction Dataset.}
% \label{tab:result_edgedet}
% \end{table}
\subsubsection{Results of Varying Exposure Manipulation.}
Compared to traditional supervised methods that learn the mapping from a poorly exposed image to a well-exposed one, our method can adjust the exposure within a certain range. To demonstrate the effectiveness of our method at varying exposure levels, we generate images with different exposures by controlling the exposure of the reference image. The results are shown in Tab.~\ref{tab:Exposure Control Combined}.
Our method performs better on the MSEC dataset compared to our dataset. This discrepancy might be attributed to the wider range of exposure values in our dataset, which varies from -2EV to +3EV, presenting a greater challenge. In contrast, the MSEC dataset has a narrower exposure range of -1.5EV to +1.5EV. In our dataset, underexposure and overexposure lead to a loss of texture, making it difficult to achieve satisfactory results solely by adjusting brightness. Additionally, in our dataset, darker images yield better results, possibly because they are closer to completely black, making conversion easier. However, overexposed images are not entirely white. The large exposure range from underexposure to overexposure further increases the challenge.
We provide the visualized results in our supplementary material.

\begin{table}[t]
\centering
\begin{subtable}[t]{\textwidth}
\centering
\begin{tabular}{@{\hskip 10pt}c@{\hskip 10pt} @{\hskip 10pt}c@{\hskip 10pt} @{\hskip 10pt}c@{\hskip 10pt} @{\hskip 10pt}c@{\hskip 10pt} @{\hskip 10pt}c@{\hskip 10pt} @{\hskip 10pt}c@{\hskip 10pt}}
\toprule
EV & -1.5 & -1.0 & 0 & +1.0 & +1.5 \\
\midrule
PSNR & 27.764 & 27.733 & 27.823 & 27.959 & 28.097 \\
\bottomrule
\end{tabular}
\caption{Exposure control on MSEC dataset.}
\label{tab:Exposure Control MSEC}
\end{subtable}
% \vspace{0.5cm} % 用于在两个子表之间添加垂直间距
\begin{subtable}[t]{\textwidth}
\centering
\begin{tabular}{@{\hskip 10pt}c@{\hskip 10pt} @{\hskip 10pt}c@{\hskip 10pt} @{\hskip 10pt}c@{\hskip 10pt} @{\hskip 10pt}c@{\hskip 10pt} @{\hskip 10pt}c@{\hskip 10pt} @{\hskip 10pt}c@{\hskip 10pt} @{\hskip 10pt}c@{\hskip 10pt}}
\toprule
EV & -2 & -1 & 0 & +1 & +2 & +3 \\
\midrule
PSNR & 22.577 & 20.528 & 18.336 & 17.820 & 15.752 & 15.138 \\
\bottomrule
\end{tabular}
\caption{Exposure control on Our dataset.}
\end{subtable}
\caption{Performance of exposure control on both datasets.}
\label{tab:Exposure Control Combined}
\end{table}

\subsubsection{Results on Edge Detection.} Exposure correction not only enhances image aesthetics but also supports downstream tasks in computer vision. Since many algorithms rely on images with standard exposure, variations in real-world exposure can challenge their effectiveness. By improving visibility and detail clarity, exposure correction enhances image features crucial for accurate object recognition. However, it is important to maintain feature integrity to avoid compromising details with artifacts. To assess its effect on crucial features, we focused on edge detection—a task highly sensitive to exposure changes—using the LDC network \cite{soria2022ldc} as a benchmark to compare performance across different exposures.

\begin{figure}[t]
    \centering
    \parbox[t][0.05\textwidth][c]{0.10\textwidth}{\scriptsize \centering Input}
    \parbox[t][0.05\textwidth][c]{0.10\textwidth}{\scriptsize \centering ECM \cite{Eyiokur2022}\\(Sup.)}
    \parbox[t][0.05\textwidth][c]{0.10\textwidth}{\scriptsize \centering Ours\\(Unsup.)}
    \parbox[t][0.05\textwidth][c]{0.10\textwidth}{\scriptsize \centering GT}
    \parbox[t][0.05\textwidth][c]{0.10\textwidth}{\scriptsize \centering Input}
    \parbox[t][0.05\textwidth][c]{0.10\textwidth}{\scriptsize \centering ECM \cite{Eyiokur2022}\\(Sup.)}
    \parbox[t][0.05\textwidth][c]{0.10\textwidth}{\scriptsize \centering Ours\\(Unsup.)}
    \parbox[t][0.05\textwidth][c]{0.10\textwidth}{\scriptsize \centering GT}
    \\
    \includegraphics[width=0.10\textwidth]{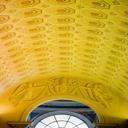}
    \includegraphics[width=0.10\textwidth]{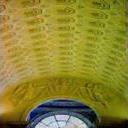}
    \includegraphics[width=0.10\textwidth]{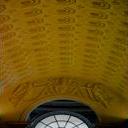}
    \includegraphics[width=0.10\textwidth]{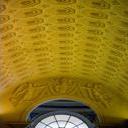}
    \includegraphics[width=0.10\textwidth]{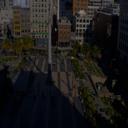}
    \includegraphics[width=0.10\textwidth]{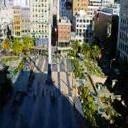}
    \includegraphics[width=0.10\textwidth]{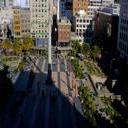}
    \includegraphics[width=0.10\textwidth]{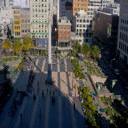}
    \\
    \includegraphics[width=0.10\textwidth]{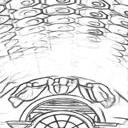}
    \includegraphics[width=0.10\textwidth]{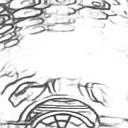}
    \includegraphics[width=0.10\textwidth]{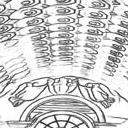}
    \includegraphics[width=0.10\textwidth]{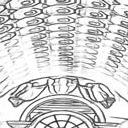}
    \includegraphics[width=0.10\textwidth]{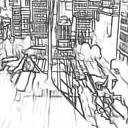}
    \includegraphics[width=0.10\textwidth]{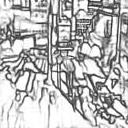}
    \includegraphics[width=0.10\textwidth]{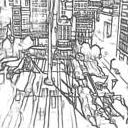}
    \includegraphics[width=0.10\textwidth]{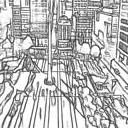}
    % \\
    % \includegraphics[width=0.10\textwidth]{img/stack/pos3/401_0_0.jpg}
    % \includegraphics[width=0.10\textwidth]{img/stack/pos3/401_0_1.jpg}
    % \includegraphics[width=0.10\textwidth]{img/stack/pos3/401_0_2.jpg}
    % \includegraphics[width=0.10\textwidth]{img/stack/pos3/401_0_3.jpg}
    % \includegraphics[width=0.10\textwidth]{img/stack/neg2/102_0_0.jpg}
    % \includegraphics[width=0.10\textwidth]{img/stack/neg2/102_0_1.jpg}
    % \includegraphics[width=0.10\textwidth]{img/stack/neg2/102_0_2.jpg}
    % \includegraphics[width=0.10\textwidth]{img/stack/neg2/102_0_3.jpg}
    % \\
    % \includegraphics[width=0.10\textwidth]{img/stack/pos3/401_1_0.jpg}
    % \includegraphics[width=0.10\textwidth]{img/stack/pos3/401_1_1.jpg}
    % \includegraphics[width=0.10\textwidth]{img/stack/pos3/401_1_2.jpg}
    % \includegraphics[width=0.10\textwidth]{img/stack/pos3/401_1_3.jpg}
    % \includegraphics[width=0.10\textwidth]{img/stack/neg2/102_1_0.jpg}
    % \includegraphics[width=0.10\textwidth]{img/stack/neg2/102_1_1.jpg}
    % \includegraphics[width=0.10\textwidth]{img/stack/neg2/102_1_2.jpg}
    % \includegraphics[width=0.10\textwidth]{img/stack/neg2/102_1_3.jpg}
    \caption{Edge detection performance by ECM \cite{Eyiokur2022} and our UEC.}
    \label{fig:edgedet}
\end{figure}

Using Radiometry Correction Dataset, we compared ECM's \cite{Eyiokur2022} method against our UEC approach. The findings, detailed in Tab.~\ref{tab:result_edgedet} and illustrated in Fig.~\ref{fig:edgedet}, highlight ECM's underperformance, even below that of unmodified inputs. This shortfall is attributed to ECM's tendency to lose fine details, negating the benefits of exposure correction. Conversely, our UEC method outperforms, demonstrated by higher PSNR values indicating superior image quality. Notably, our unsupervised approach excels in adapting to varied exposures for diverse computer vision tasks, requiring only the preservation of RAW files during data collection.

\subsubsection{Impact of Reference Images on UEC Results.} 
\begin{figure}[t]
    \centering
    \parbox[t][1em][c]{0.09\textwidth}{\tiny \centering Input}
    \parbox[t][1em][c]{0.09\textwidth}{\tiny \centering Ref. Image1}
    \parbox[t][1em][c]{0.09\textwidth}{\tiny \centering Result1}
    \parbox[t][1em][c]{0.09\textwidth}{\tiny \centering Ref. Image2}
    \parbox[t][1em][c]{0.09\textwidth}{\tiny \centering Result2}
    \hspace{0.1em}
    \parbox[t][1em][c]{0.09\textwidth}{\tiny \centering Input}
    \parbox[t][1em][c]{0.09\textwidth}{\tiny \centering Ref. Image1}
    \parbox[t][1em][c]{0.09\textwidth}{\tiny \centering Result1}
    \parbox[t][1em][c]{0.09\textwidth}{\tiny \centering Ref. Image2}
    \parbox[t][1em][c]{0.09\textwidth}{\tiny \centering Result2}
    \\
    \centering
    \includegraphics[width=0.09\textwidth]{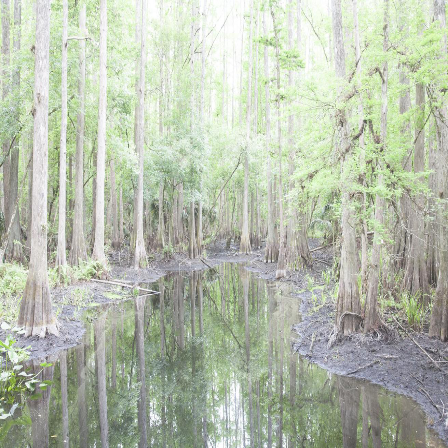}
    \includegraphics[width=0.09\textwidth]{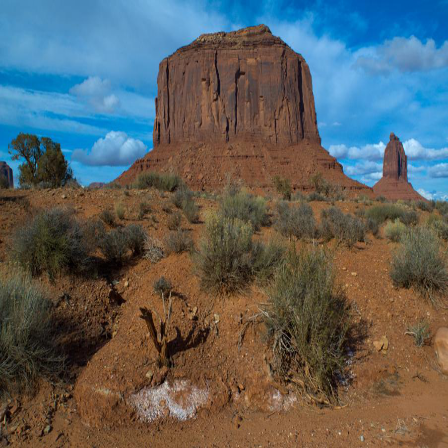}
    \includegraphics[width=0.09\textwidth]{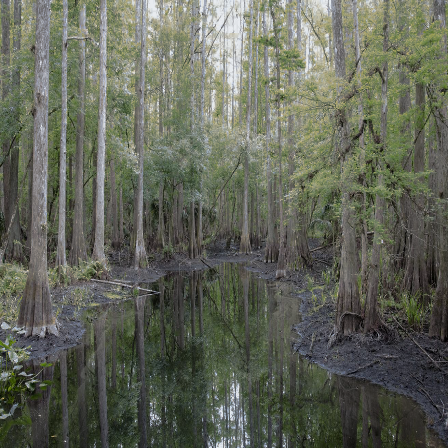}
    \includegraphics[width=0.09\textwidth]{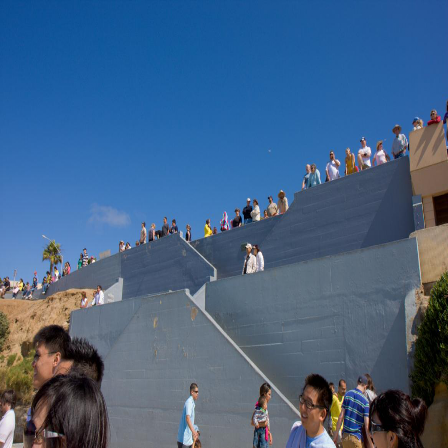}
    \includegraphics[width=0.09\textwidth]{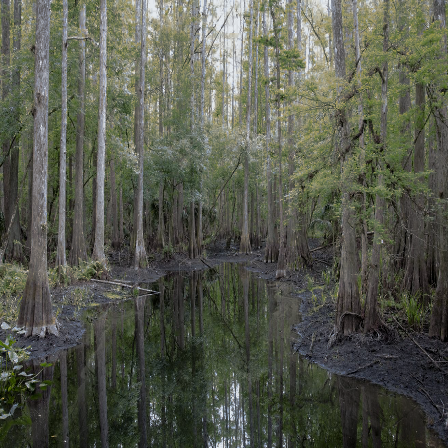}
    \hspace{0.1em}
    \includegraphics[width=0.09\textwidth]{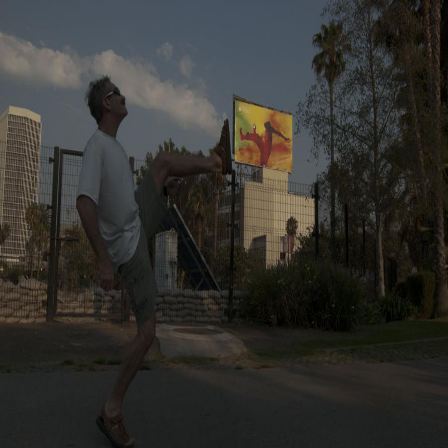}
    \includegraphics[width=0.09\textwidth]{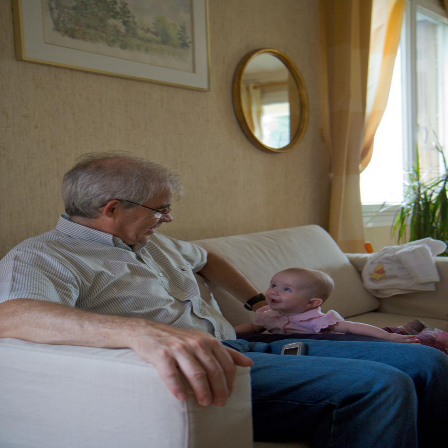}
    \includegraphics[width=0.09\textwidth]{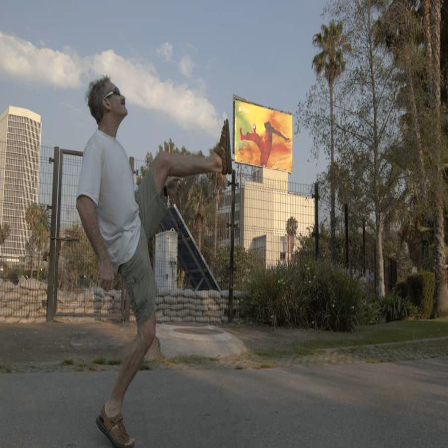}
    \includegraphics[width=0.09\textwidth]{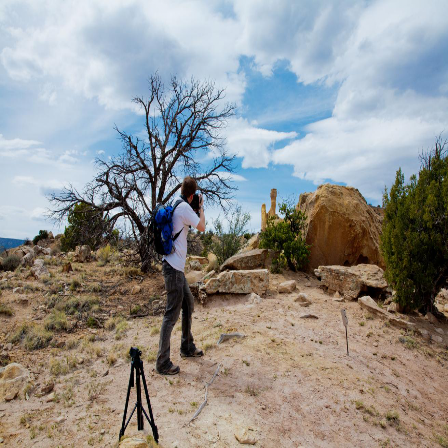}
    \includegraphics[width=0.09\textwidth]{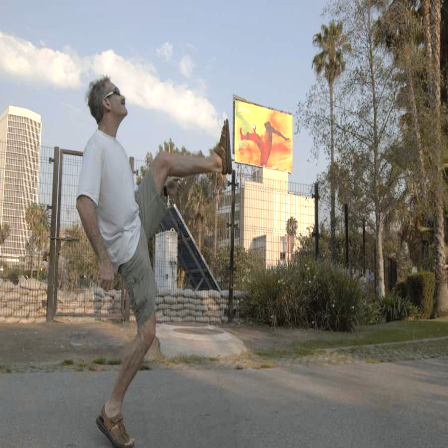}
    \caption{Impact of Reference Images on UEC Results.}
    \label{fig:ref_imgs}
\end{figure}

Fig.~\ref{fig:ref_imgs} demonstrates the impact of utilizing different reference images on the outcomes of our UEC model. Notably, the results show minor variations within a certain range, while maintaining their overall performance. This observation emphasizes the UEC model's robust generalizability, its proficiency in extracting exposure features from diverse images, and its competence in achieving exposure alignment across a range of scenes.

% \begin{table}[ht]
%     \centering
%     \begin{tabular}{p{1.5cm}cccc}
%         \toprule
%         Method & Parameters & Time(GPU) & Time(CPU) \\
%         &(K) & (ms) & (ms) \\
%         \midrule
%         ECM \cite{Eyiokur2022} & 182,443 & 7.08 & 149.39  \\
%         Ours & 19 & 1.46 & 6.38  \\
%         \bottomrule
%     \end{tabular}
%     \caption{Efficiency comparison at a resolution of 256$\times$256.}
%     \label{tab:speed1}
% \end{table}

% \subsection{Time and Memory Consumption}\label{sec:time}
% Remarkably, our approach achieves a level of performance comparable to ECM \cite{Eyiokur2022} while utilizing merely 0.01\% of their parameters. Our model is capable of real-time operation at a 4K resolution and can directly process high-definition images, mitigating any low-level feature loss incurred due to downsampling. To assess the inference speed, we conducted experiments using 256$\times$256 images on both an Intel(R) Xeon(R) Gold 6248 CPU @ 2.50GHz and a NVIDIA TESLA A100 GPU. In a rigorous evaluation, each model underwent 1000 runs under identical conditions to accurately compute the average runtime. The results is shown in Tab.~\ref{tab:speed1}. Specifically, our method achieves approximately 4.85 times faster inference on the GPU and 23.41 times faster inference on the CPU compared to ECM \cite{Eyiokur2022}. This performance advantage underscores the practical viability of our approach in real-time applications and resource-constrained scenarios. Comprehensive results can be found in the supplementary material.

\subsection{Time and Memory Consumption}\label{sec:time}

Our approach achieves comparable performance to ECM \cite{Eyiokur2022} with only 0.01\% of their parameters. It supports real-time operation at 4K resolution, processing high-definition images directly and avoiding low-level feature loss from downsampling. As shown in Tab.~\ref{tab:speed1}, our method is 4.85 times faster on GPU and 23.41 times faster on CPU compared to ECM \cite{Eyiokur2022}. Additional results can be found in the supplementary material.

\begin{table*}[t]
    \centering
    \begin{tabular}{@{\hskip 10pt}c@{\hskip 10pt} @{\hskip 10pt}c@{\hskip 10pt} @{\hskip 10pt}c@{\hskip 10pt} @{\hskip 10pt}c@{\hskip 10pt} @{\hskip 10pt}c@{\hskip 10pt}}
    \toprule
    Method & Parameters & Model Size & Time(GPU) & Time(CPU) \\
    \midrule
    ECM. \cite{Eyiokur2022} & 182,443,267 & 695 MB & 7.08ms & 149.39ms  \\
    ours & 19,388 & 0.079 MB & 1.46ms & 6.38ms  \\
    \bottomrule
    \end{tabular}
    \caption{Comparison at a resolution of 256$\times$256 between our UEC model and ECM~\cite{Eyiokur2022}.}
    \label{tab:speed1}
\end{table*}

% Our approach achieves comparable performance to ECM \cite{Eyiokur2022} with only 0.01\% of their parameters. It supports real-time operation at 4K resolution, processing high-definition images directly and avoiding low-level feature loss from downsampling. We tested inference speed with 256$\times$256 images on an Intel(R) Xeon(R) Gold 6248 CPU @ 2.50GHz and an NVIDIA TESLA A100 GPU, running each model 1000 times to calculate average runtime. As shown in the supplementary material, our method is 4.85 times faster on GPU and 23.41 times faster on CPU compared to ECM \cite{Eyiokur2022}, highlighting its real-time application viability in resource-constrained scenarios.

% \begin{table}[ht]
%     \centering
%     \begin{tabular}{c c c c c}
%     \toprule
%     Resolution & Width & Height & Time(GPU) & Time(CPU) \\
%     \midrule
%     4K & 4096 & 2160 & 23.32ms & N/A  \\
%     2K & 2560 & 1440 & 9.40ms & N/A  \\
%     1080P & 1920 & 1080 & 5.49ms & 476.08ms \\
%     720P & 1280 & 720 & 2.72ms & 212.08ms \\
%     480P & 720 & 480 & 1.44ms & 81.11ms \\
%     \bottomrule
%     \end{tabular}
%     \caption{Comparison on different resolutions for our method}
%     \label{tab:speed2}
% \end{table}

\section{Conclusion}
In this study, we introduce an unsupervised methodology for correcting exposure, overcoming three primary limitations observed in prior methods: the dependence on labor-intensive paired datasets, constrained generalizability, and the deterioration of low-level features. To address these issues, our approach unfolds in three stages. Initially, we establish an unsupervised learning framework grounded in the inherent principles of exposure, thereby obviating the requirement for ground truth data. Subsequently, we eschew the traditional reliance on ground truth, which often incorporates subjective stylistic biases, and instead, draw insights from a diverse array of images differing solely in their radiometric properties. Finally, we employ color transformation techniques to maintain the intrinsic pixel relationships, effectively preserving low-level features and enhancing the robustness of our method.

\bibliographystyle{splncs04}
\bibliography{egbib}
\end{document}